\documentclass[lettersize,journal]{IEEEtran}
\usepackage[utf8]{inputenc}
\usepackage[T1]{fontenc}
\usepackage{amsmath,amsfonts,amssymb,mathtools,amsthm}
\usepackage{algorithmic}
\usepackage{algorithm}
\usepackage{array}
\usepackage[caption=false,font=normalsize,labelfont=sf,textfont=sf]{subfig}
\usepackage{textcomp}
\usepackage{stfloats}
\usepackage{url}
\usepackage{verbatim}
\usepackage{graphicx}
\usepackage{cite}
\usepackage{tabularx}
\usepackage{booktabs,multirow,makecell}
\usepackage[table]{xcolor}
\usepackage{microtype,nicefrac,pifont}
\usepackage{paralist,comment}
\usepackage[hidelinks]{hyperref}
\graphicspath{{./}}
\newcommand{\cmark}{\ding{51}}
\newcommand{\xmark}{\ding{55}}
\newcommand{\textBF}[1]{\textbf{#1}}
\definecolor{Gray}{gray}{0.9}
\theoremstyle{plain}

\theoremstyle{definition}

\theoremstyle{remark}

\newcommand{\norm}[1]{\left\lVert#1\right\rVert}

\begin{document}

\title{Robust Dynamic Expansion for Continual Learning under Backdoor Attacks via Purification and Selective Recovery}

\author{Keyu~Lin,
        Fei~Ye*,
        Qihe~Liu,
        Shijie~Zhou,
        and~Jiguo~Yu,~\IEEEmembership{Fellow,~IEEE}%
\thanks{
Keyu Lin, Fei Ye, Qihe Liu, Shijie Zhou, and Jiguo Yu
are with the University of Electronic Science and Technology
of China, Chengdu 610054, China
(e-mail: 202522090409@std.uestc.edu.cn;
feiye@uestc.edu.cn;
qiheliu@uestc.edu.cn;
sjzhou@uestc.edu.cn;
jiguoyu17@uestc.edu.cn).%
}
\thanks{* Corresponding author.}%
}

\maketitle

\begin{abstract}
Continual learning (CL) enables models to acquire new knowledge from sequentially arriving tasks while retaining previously learned knowledge. However, in practical scenarios, task streams collected from untrusted sources may contain backdoor-poisoned samples, posing a critical challenge to the stability, plasticity, and security of continual learners. In this work, we investigate a challenging setting termed Continual Learning Under Backdoor Attack (CLUBA), where each incremental task may involve a small proportion of maliciously manipulated training samples. Unlike conventional continual learning or backdoor defense scenarios, CLUBA requires models to simultaneously mitigate catastrophic forgetting, preserve adaptation capability, and prevent the absorption of malicious supervision during sequential updates. To address this challenge, we propose a robust dynamic-expansion framework that integrates sample purification, selective recovery, and robust expert routing into a unified continual learning paradigm. Specifically, we introduce Bi-Prototype Purification (BPP) to identify suspicious samples by exploiting semantic discrepancies in feature space. Based on purified data, Gradient Discrepancy-based Robustness Optimization (GDBRO) selectively recovers informative poisoned samples through pseudo-label correction and gradient consistency evaluation, improving robustness while maintaining model plasticity. Furthermore, Robust Feature Consistency-based Expert Selection (RFCBES) constructs perturbation-aware class prototypes to enable reliable expert routing under corrupted or shifted inputs.
\end{abstract}


\section{Introduction}

\IEEEPARstart{C}{ontinual} learning (CL) corresponds to a training paradigm in which the model is sequentially exposed to discrete tasks, with access restricted to only a small subset of samples belonging to the current task, while all previous training data become unavailable. Optimization under this paradigm typically induces significant performance degradation on previously learned tasks, a well-documented phenomenon termed catastrophic forgetting \cite{cl_survey2}. This degradation stems from the tendency of the standard training process to overwrite model parameters wholesale in adapting to new task data. To mitigate these issues, numerous methodologies have been developed, broadly categorized into three families: regularization-based methods \cite{EWC}, which impose constraints on updates to parameters deemed important for prior tasks; memory replay approaches \cite{RainbowMemory}, which leverage a fixed-capacity buffer of exemplars to preserve knowledge through rehearsal; and dynamic network architectures \cite{pnn}, which expand network capacity by instantiating new parameters for learning emerging tasks.

In practical applications, continual learners are often updated with continuously arriving data collected from heterogeneous and potentially unreliable sources \cite{air}. Such data streams may contain maliciously manipulated samples designed to implant hidden behaviors into the model. This motivates a more challenging setting, referred to as Continual Learning Under Backdoor Attack (CLUBA), where each task may contain a small proportion of backdoor-poisoned training samples. Compared with conventional continual learning, CLUBA introduces an additional security constraint: the learner must not only preserve previous knowledge and adapt to new tasks, but also avoid absorbing malicious supervision during incremental updates.

The difficulty of CLUBA arises from the strong coupling among three objectives. First, the model must maintain stability to prevent catastrophic forgetting of previous tasks. Second, it must retain sufficient plasticity to learn new task-specific knowledge from limited current-task data. Third, it must achieve robustness against poisoned supervision, which may otherwise induce incorrect feature representations and trigger–label associations. These objectives are inherently conflicting: aggressive adaptation may amplify malicious patterns, while conservative optimization may suppress useful task information. Therefore, directly applying conventional continual learning or backdoor defense methods often leads to unsatisfactory performance under sequential poisoning scenarios.

To mitigate the threat posed by poisoning examples within CL, an early defense mechanism, termed the Anisotropic and Isotropic Replay (AIR) \cite{air}, incorporates a strategic replay buffer. This mechanism combines three distinct replay paradigms with a consistency regularization loss term to counteract catastrophic forgetting and improve model robustness. Another approach examines differential logit calibration between historical and newly encountered tasks to regulate gradient dynamics and introduces a gradient-selection criterion for preserving critical exemplars into the memory buffer \cite{aflc}. Nevertheless, these methods primarily address adversarial robustness in continual learning; their objectives do not directly provide the task-local identification and selective recovery of backdoor-poisoned samples considered here. Furthermore, these methodologies typically utilize a fixed-capacity replay buffer for storing exemplars extracted from previous tasks. Such a design requires retaining historical training examples and demonstrates constrained effectiveness in managing extended sequences of tasks \cite{LifelongDynamicTS}.

In this paper, we propose a robust dynamic-expansion framework for CLUBA based on a principle: corrupted samples should not be treated as uniformly unusable; instead, they should be identified, filtered, and selectively recovered whenever their supervision can be corrected reliably. Building on this principle, our framework combines three components that address the three central challenges of CLUBA in a coordinated way. (1) Bi-Prototype Purification (BPP) separates each class into candidate clean and candidate poisoned subsets in semantic feature space, exploiting the observation that poisoned samples tend to deviate from the dominant semantic structure of their class.
(2) Gradient Discrepancy-based Robustness Optimization (GDBRO) first warms up the current expert on purified clean data to recover plasticity, and then selectively reincorporates reliable candidate poisoned samples through pseudo-label correction and gradient-based filtering to improve robustness without amplifying label noise.
(3) Robust Feature Consistency-based Expert Selection (RFCBES) constructs perturbation-aware class prototypes for inference-time routing, making expert selection more reliable under corrupted or shifted inputs.

Our method differs from prior continual learning approaches in two key aspects. First, rather than viewing robustness as an auxiliary add-on to anti-forgetting mechanisms, we treat robustness and continual adaptation as a joint optimization problem. Second, instead of relying on replay buffers to compensate for corrupted updates, we leverage task-local purification and selective recovery to improve both robustness and plasticity while preserving the benefits of dynamic expansion. We evaluate the proposed framework on multiple continual backdoor benchmarks under both clean and poisoned settings. The results show that our method consistently outperforms strong baselines, achieving better robustness on current tasks while maintaining competitive retention on previous ones. The contributions of this paper are summarized as~: \begin{itemize}[(1)]
\item We study CLUBA, a continual learning setting in which each task may contain a small fraction of poisoned training samples, and highlight the joint challenge of forgetting, weak robustness, and reduced plasticity;
\item We propose a unified robust dynamic-expansion framework that couples purification, selective poisoned-sample recovery, and robust expert routing, rather than treating these goals independently;
\item We demonstrate through extensive experiments that the proposed framework improves both adaptation and retention under task-wise backdoor attacks, outperforming strong continual learning baselines across multiple benchmarks.
\end{itemize}

\section{Related Work}

\noindent
\textBF{Continual Learning.}
Continual Learning (CL) addresses catastrophic forgetting when learning from a task stream~\cite{cl_survey,cl_survey2}. Regularization-based methods such as EWC~\cite{EWC}, SI~\cite{si}, MAS~\cite{mas}, and LwF~\cite{lwf} constrain important parameters or preserve historical predictions. Replay-based methods~\cite{icarl,der,er,mmir,er_ace} rehearse stored examples, whereas gradient-based methods such as GEM and A-GEM project new-task gradients to reduce interference with previously acquired knowledge~\cite{gem,agem}. Prompt-based approaches, including L2P~\cite{l2p}, DualPrompt~\cite{dualprompt}, CODA-Prompt~\cite{coda}, and HiP~\cite{hip}, as well as parameter-efficient adaptation methods~\cite{moe,pet}, reduce the number of parameters updated for each new task.

Recent pre-trained-model-based approaches further improve the stability--plasticity trade-off by integrating representations from multiple pre-trained backbones and adaptively controlling the optimization of individual representation layers~\cite{lms}. When historical samples cannot be retained because of privacy or storage constraints, data-free knowledge distillation provides an alternative means of transferring prior knowledge~\cite{vd}. Continual-learning techniques have also been extended to generative settings, where task-aware memory enhancement and elastic-concept distillation preserve previously learned concepts during lifelong text-to-image generation~\cite{cyw}. Nevertheless, these methods generally assume that the incoming training data are benign and do not explicitly address backdoor-corrupted task streams.

\noindent
\textBF{Backdoor Attacks and Defenses.} Backdoor attacks inject hidden triggers (BadNets~\cite{badnets}, Blend~\cite{blend}, WaNet~\cite{wanet}, ISSBA~\cite{issba}, Clean-Label~\cite{clean_label}, BadToken~\cite{badtoken}). In CL, models are uniquely vulnerable to sequential data poisoning~\cite{poison_cl1, poison_cl2}. Since traditional backdoors suffer from forgetting, recent works motivate persistent attacks like LTB~\cite{persistent_backdoor} (targeting stable neurons) and CBACL~\cite{cbacl} (addressing label shift via anchor features). Defenses include fine-tuning~\cite{ft}, Neural Cleanse~\cite{neural_cleanse}, Fine-Pruning~\cite{fine_pruning}, STRIP~\cite{strip}, ABL~\cite{abl}, AC~\cite{ac}, NAD~\cite{nad}, ANP~\cite{anp}, I-BAU~\cite{i_bau}, and adversarial inspection~\cite{adversarial, dbd}. However, most static defenses fail against dynamic CL threats.

\begin{figure*}[t]
    \centering
    \includegraphics[width=0.99\linewidth]{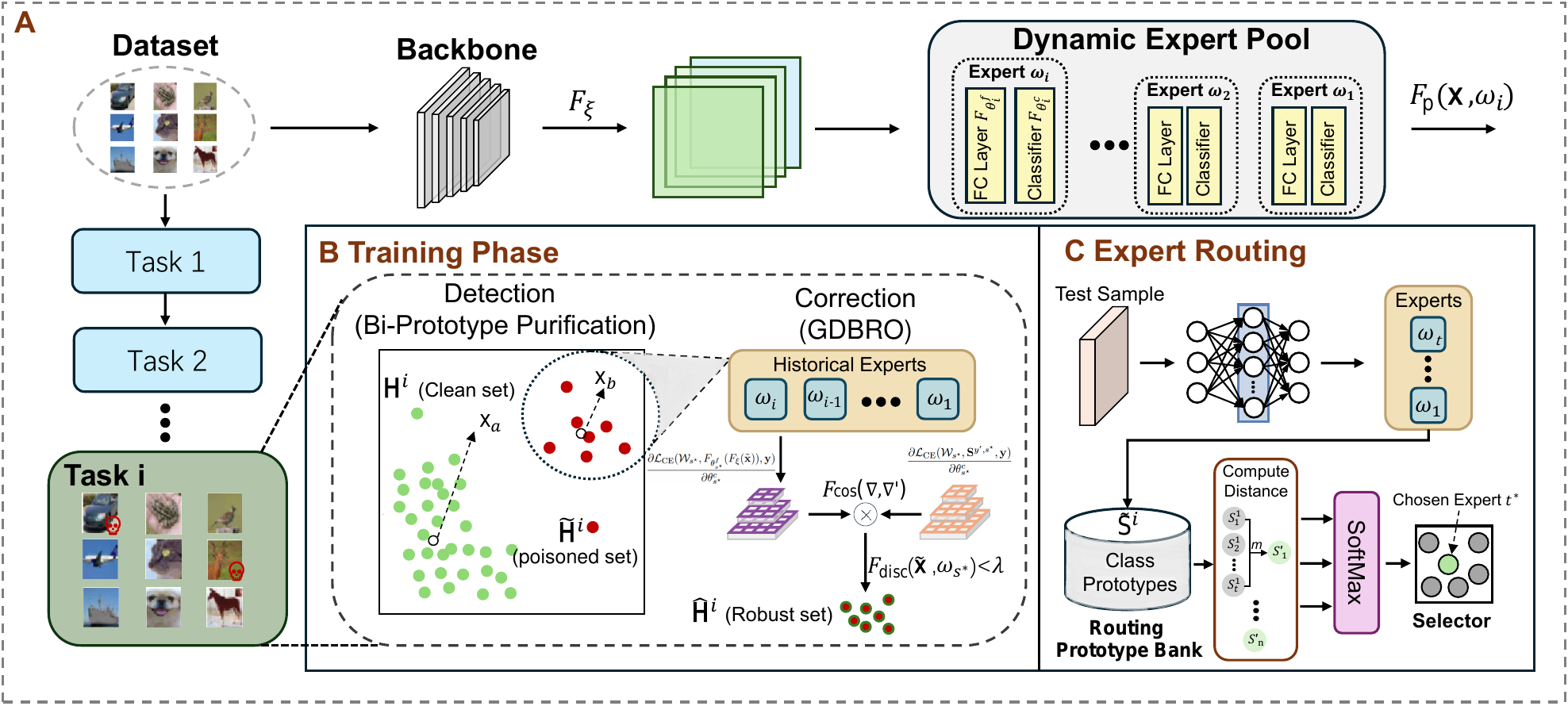}
\vspace{-5pt}
    \caption{Overview of our proposed continual data backdoor defense framework. \textbf{(A) Overall Architecture:} A shared backbone extracts features for sequential tasks, utilizing a Dynamic Expert Pool to prevent catastrophic forgetting. \textbf{(B) Training Phase:} The Bi-Prototype Purification (BPP) mechanism detects candidate poisoned samples. The proposed GDBRO module then rescues clean samples by evaluating gradient similarities against all experts, forming a robust training set. \textbf{(C) Expert Routing:} During inference, test samples are accurately routed to the optimal task-specific expert based on their feature distances to the robust class prototypes stored in the Routing Prototype Bank.}
    \vspace{-10pt}
\label{fig:overall_architecture}
\end{figure*}

\noindent
\textBF{Dynamic Networks for Continual Learning.}
Architecture-based continual-learning methods isolate task knowledge through dynamic network expansion~\cite{dem_survey,pnn,packnet}. Representative methods such as DEM~\cite{dem} and DyTox~\cite{dytox} preserve previously learned components while allocating additional capacity to new tasks, whereas mixture-of-experts adapters~\cite{moe} improve parameter efficiency. More recently, the Adaptive and Expandable Mixture Model employs invariant and evolving representation networks and incrementally introduces lightweight expert modules, thereby balancing stable knowledge preservation with adaptive representation learning~\cite{lae}. The Dynamic Siamese Expansion Framework similarly combines static and dynamic backbones, but additionally optimizes the resulting representations to improve robustness against adversarial perturbations~\cite{dse}.

Despite these advantages, dynamic architectures remain vulnerable to training-time backdoor attacks. A newly allocated expert may directly learn trigger--label associations from its current task, while freezing that expert can cause the acquired backdoor to persist throughout subsequent tasks. Moreover, existing robust dynamic-expansion methods primarily consider adversarial perturbations rather than task-wise backdoor poisoning. Consequently, systematically securing dynamic experts against poisoned continual-learning streams remains underexplored.

\noindent
\textBF{Security Implications of Continual Learning Strategies.}
The security behavior of a continual learner is closely related to the mechanism used to preserve previous knowledge. Regularization-based methods constrain the modification of important parameters or historical predictions; however, the same constraints may also preserve malicious behavior if a backdoor has already been incorporated into those parameters. Replay-based methods retain representative historical samples to mitigate forgetting, but a contaminated memory may repeatedly expose the model to poisoned examples and reinforce trigger--label associations. Similarly, parameter-isolation and dynamic-expansion methods prevent later tasks from overwriting earlier experts, but this isolation can also make an implanted backdoor difficult to remove once the corresponding expert has been frozen.

Security-oriented continual learning has also been investigated in cross-domain face anti-spoofing, where rehearsal-free adapters and attention-consistency regularization reduce forgetting while improving generalization to previously unseen spoofing domains~\cite{rf}. This setting demonstrates the importance of jointly considering knowledge retention and security-related generalization. Nevertheless, face anti-spoofing focuses on recognizing presentation attacks and domain shifts at inference time, rather than identifying poisoned training samples that implant hidden behavior during continual learning.

\noindent
\textBF{Backdoor Defense in the Continual Setting.}
Defenses can intervene at data purification, robust training, or model repair. Activation Clustering filters training data in feature space; Anti-Backdoor Learning changes the training objective to suppress trigger memorization; Neural Cleanse and Fine-Pruning examine or repair a trained model~\cite{ac,abl,neural_cleanse,fine_pruning}. Applying these strategies sequentially requires accounting for task-local data availability and the preservation of earlier benign knowledge. Our framework connects purification with robust training for each arriving task without requiring access to historical images.

The resulting security problem couples stability, plasticity, and purification. Preserving learned functions can also preserve an already acquired backdoor, whereas aggressive updates intended to remove a backdoor can disturb benign knowledge. Adversarial continual-learning methods such as AFLC and AIR address robustness objectives that do not directly identify and selectively recover the backdoor-poisoned samples considered here. This motivates combining task-local data selection with isolated experts and a refined routing bank.

\section{Methodology}

\subsection{Problem Definition and Threat Model}

\noindent
\textBF{The learning scenario.} Continual learning seeks to optimize a model that can incrementally acquire new information while retaining previously acquired knowledge. Given a sequence of $n$ tasks $\{ {\mathcal{E}}_1,\cdots,{\mathcal{E}}_n \}$, each task ${\mathcal{E}}_i$ comprises a corresponding training dataset ${\mathcal{D}}_i = \{ {\bf x}_j, {\bf y}_j \}^{n_i}_{j=1}$, where $n_i$ signifies the number of training samples within ${\mathcal{D}}_i$. Here, ${\bf x}_j \in {\mathcal{X}}$ and ${\bf y}_j \in {\mathcal{Y}}$ denote the $j$-th training sample and its associated class label belonging to the data space ${\mathcal{X}}$ and label space ${\mathcal{Y}}$, respectively. In the CLUBA scenario, a portion of the training dataset for each task ${\mathcal{E}}_i$ is corrupted according to a specified threat model. During the learning phase for each task ${\mathcal{E}}_i$, the victim model has access exclusively to the current training dataset ${\mathcal{D}}_i$, while all prior datasets $\{ {\mathcal{D}}_1,\cdots,{\mathcal{D}}_{i-1} \}$ remain unavailable.

\noindent \textBF{The threat model.} The CLUBA scenario formulates each threat model as a data poisoning mechanism designed to corrupt a small subset of training samples. In this study, we consider implementing the data poisoning mechanism using the backdoor attack algorithm \cite{backdoor_survey}. The defender receives the potentially poisoned task data and knows the permitted class set of the current training task. Attacker access depends on the attack configuration described in the experimental protocol. Specifically, for a given training dataset $\mathcal{D}_i$, a small, randomly selected subset $\tilde{\mathcal{D}}_i$ is drawn from $\mathcal{D}_i$. Each clean sample ${\bf x}$ within $\tilde{\mathcal{D}}_i$ is then adversarially corrupted by its corresponding threat model via the transformation $\tilde{\bf x} = M_i( {\bf x} )$. The resulting contaminated samples may include input perturbations, label corruption, or a combination thereof, contingent upon the specified threat model. This setting differs from standard continual learning in that the learner must simultaneously (i) retain knowledge from previous tasks, (ii) adapt to the current task, and (iii) avoid overfitting to corrupted supervision within the current task.

\subsection{Dynamic Expansion Model}

Within the domain of continual learning, the Dynamic Expansion Model (DEM) constitutes a potent strategy for mitigating forgetting. This architectural paradigm autonomously generates a new expert module for each subsequent task, while existing model parameters remain frozen to conserve previously acquired knowledge. By leveraging Pretrained Models (PTMs) as its foundational backbone, DEM augments performance in sequential task scenarios. Building upon its established efficacy, this work investigates a PTM-based dynamic expansion framework to address the problem of Continual Learning Under Backdoor Attack (CLUBA). Formally, let $F_{\xi}: \mathcal{X} \to \mathcal{Z}$ denote a primary representation network instantiated by a PTM. This network maps an input sample $\mathbf{x}$ to a representation $\mathbf{z} \in \mathcal{Z}$ within the feature space $\mathcal{Z}$. To encapsulate task-specific information, an expert module $\mathcal{W}_i$ is defined, comprising a feature transformation function $F_{\theta^f_i}: \mathcal{Z} \to \mathcal{{Z}}'$ and a linear classifier $F_{\theta^c_i}: \mathcal Z\times\mathcal Z' \to \mathcal{Y}$, where subscript $i$ indexes the expert. The transformation function $F_{\theta^f_i}$ processes the output from the primary backbone $F_{\xi}$ to produce a task-specific representation $\mathbf{{\bf z}}' \in \mathcal{{Z}}'$, where $\mathcal{{Z}}'$ denotes the feature space. This representation is concatenated with the shared backbone feature and subsequently passed to the linear classifier $F_{\theta^c_i}$ to yield a prediction~:
\begin{equation}
\begin{aligned}
F_{\rm p}( {\bf x}, {\mathcal{W}}_i ) = F_{\theta^c_i}([F_{\xi}({\bf x});F_{\theta^f_i}(F_{\xi}({\bf x}))])\,,
\end{aligned}
\end{equation}
\noindent where ${y}' = F_{\rm p}( {\bf x}, {\mathcal{W}}_i )$ is the predicted probability vector.

For clarity, define the classifier input and normalized shared feature as
\begin{equation}
\begin{aligned}
h_i(x)&=[F_\xi(x);F_{\theta_i^f}(F_\xi(x))],\\
u(x)&=F_\xi(x)/\norm{F_\xi(x)}_2.
\end{aligned}\label{eq:representations}
\end{equation}
The shared ViT backbone remains frozen. BPP and expert routing use $u(x)$; classification and sample-gradient evaluation use $h_i(x)$.

\subsection{Backdoor Sample Detecting via Bi-Prototype Purification Mechanism}

Poisoned samples often differ from clean samples only through subtle perturbations in image or label space, making them difficult to detect directly at the pixel level. We therefore identify candidate poisoned samples in semantic feature space, where class-level structure is more stable. To achieve this goal, we introduce a novel Bi-Prototype Purification (BPP) mechanism to identify backdoor samples by assessing knowledge similarity within the semantic feature space rather than the original image space. Specifically, when learning the $i$-th task ${\mathcal{E}}_i$, we obtain the training dataset ${\mathcal{D}}_i$, which may contain a small proportion of backdoor samples. The BPP mechanism first partitions ${\mathcal{D}}_i$ into ${c}_i$ subsets $\{{\bf C}^1_i, \cdots, {\bf C}^{c_i}_i \}$ based on category labels. Each subset, defined as ${\bf C}^{j}_i = F_{\rm split}({\mathcal{D}}_i, {\mathcal{C}}^{j}_i)$, comprises samples belonging to the same class~:
\begin{equation}
\begin{aligned}
&F_{\rm split}({\mathcal{D}}_i,{\mathcal{C}}^j_i) \\
&\quad=\{ {\bf x}_s\mid F_{\rm cat}({\bf x}_s)={\mathcal{C}}^j_i,\;{\bf x}_s\in{\mathcal{D}}_i,\\
&\hspace{38pt}s=1,\ldots,|{\mathcal{D}}_i|\}\,.
\end{aligned}
\end{equation}
\noindent where the symbol $\mathcal{C}^j_i$ denotes the classification label associated with cluster $\mathbf{C}^j_i$, while the function $F_{\mathrm{cat}}(\cdot)$ provides the observed training label for any given data sample $\mathbf{x}_s$. The cardinality $|\mathcal{D}_i|$ represents the total sample count in dataset $\mathcal{D}_i$. As each class subset may contain backdoor instances, BPP separates candidate clean and suspicious data using two-center clustering of normalized shared features. Samples whose observed labels fall outside the current task's permitted class set are first assigned to the suspicious set. For each remaining class, the clustering objective is
\begin{equation}
\min_{\mu_a,\mu_b,z_s\in\{a,b\}}
\sum_{x_s\in\mathbf C_i^j}\norm{u(x_s)-\mu_{z_s}}_2^2.
\label{eq:kmeans}
\end{equation}
The resulting groups $\mathbf H_a^{j,i}$ and $\mathbf H_b^{j,i}$ contain samples assigned to their respective centers. Define their mean within-cluster distances and the smaller-cluster fraction by
\begin{equation}
\begin{aligned}
d_k&=\frac{1}{|\mathbf H_k^{j,i}|}\sum_{x\in\mathbf H_k^{j,i}}\norm{u(x)-\mu_k}_2,\\
r&=\frac{\min(|\mathbf H_a^{j,i}|,|\mathbf H_b^{j,i}|)}{|\mathbf C_i^j|}.
\end{aligned}\label{eq:cluster_stats}
\end{equation}
BPP exploits the observation that a shared trigger may induce a compact subcluster, whereas benign examples exhibit greater semantic diversity. To avoid forcing a split in balanced, naturally multimodal classes, the whole class is retained when $r>0.35$ and its sample count is at most 1.5 times the median count over present valid classes. Otherwise, the more dispersed group is retained as candidate clean data and the more compact group is marked suspicious. Classes with fewer than ten examples or degenerate clustering outcomes are retained without splitting.

For notational brevity, $\mathbf H^{j,i}$ and $\tilde{\mathbf H}^{j,i}$ denote the candidate clean and suspicious subsets for class $j$ in task $i$. Their unions form $\mathbf H^i$ and $\tilde{\mathbf H}^i$. These are selection decisions rather than ground-truth clean/poison labels; GDBRO subsequently evaluates suspicious samples for selective recovery.

\noindent
\textBF{Implementation and fallback conventions.}
The two-center K-Means procedure uses ten initializations and random state 42. If the within-cluster dispersions are tied, cluster 1 is retained. A clustering exception or a single occupied cluster causes the whole valid class to be retained. Together with the small-class and balance/count checks described above, these conventions avoid imposing a clean/poison split when the clustering result is uninformative.

\begin{table*}[t]
    \centering
    \caption{Basic image configurations and preprocessing transformations.}
    \vspace{-5pt}\label{tab:dataset_transformations}
        \setlength{\tabcolsep}{20pt}
        \begin{tabular}{l c l l}
            \toprule
            \textbf{Dataset} & \textbf{Original Size} & \textbf{Phase} & \textbf{Resizing \& Augmentation} \\
            \midrule
            \multirow{2}{*}{Split CIFAR-10} & \multirow{2}{*}{$3 \times 32 \times 32$} & Training & Random crop (pad 4), Horizontal Flip, Resize $224 \times 224$ \\
             & & Testing & Resize $224 \times 224$ \\
            \midrule
            \multirow{2}{*}{Split CIFAR-100} & \multirow{2}{*}{$3 \times 32 \times 32$} & Training & Random crop (pad 4), Horizontal Flip, Resize $224 \times 224$ \\
             & & Testing & Resize $224 \times 224$ \\
            \midrule
            \multirow{2}{*}{Split Tiny-ImageNet} & \multirow{2}{*}{$3 \times 64 \times 64$} & Training & Random crop (pad 4), Horizontal Flip, Resize $224 \times 224$ \\
             & & Testing & Resize $224 \times 224$ \\
            \bottomrule
        \end{tabular}
        \vspace{-10pt}
\end{table*}

\subsection{Gradient Discrepancy-based Robustness Optimization}

Training only on the purified clean set ${\bf H}^i$ can protect the active expert from corrupted supervision, but it also discards potentially useful information contained in the candidate poisoned subset. To balance robustness and plasticity, we introduce Gradient Discrepancy-based Robustness Optimization (GDBRO) to correct backdoor samples and integrate them during the training process. Specifically, the proposed GDBRO mechanism consists of a warm-up optimization process to enhance the model's discrimination ability and an integration optimization process to selectively integrate many critical training samples for improving model robustness.  

\noindent
\textBF{The warm-up optimization process.} This strategy aims to optimize the current expert module ${\mathcal{W}}_i$ on the clean training dataset ${\bf H}^i$ during the $i$-th task learning. This is implemented by using the cross-entropy loss, expressed as~:
\begin{equation}
\begin{aligned}
&{\mathcal{L}}_{\rm warm}\\
&={\mathbb E}_{({\bf x},{\bf y})\sim P_{{\bf H}^i}}
\left[-\sum_{j=1}^{|{\mathcal{Y}}|}{\bf y}[j]\log F_p({\bf x},{\mathcal{W}}_i)[j]\right].
\end{aligned}
\end{equation}
\noindent where $P_{{\bf H}^i}$ denotes the distribution of the set ${\bf H}^i$ and $|{\mathcal{Y}}|$ represents the total number of categories. ${\bf y}[j]$ and $F_{p}( {\bf x}, {\mathcal{W}}_i )[j]$ denote the $j$-th dimension of the class label ${\bf y}$ and the predicted vector $F_{p}( {\bf x}, {\mathcal{W}}_i )$, respectively. Once the training process of the current active expert ${\mathcal{W}}_i$ is finished, the proposed GDBRO mechanism captures the statistical information for each category by determining the class prototype, which can be used for the backdoor sample correction process. Specifically, for the given set $\mathbf H^{j,i}$, the class prototype is the normalized mean of normalized shared features:
\begin{equation}
\mathbf S^{j,i}=\operatorname{normalize}\left(
\frac{1}{|\mathbf H^{j,i}|}\sum_{x\in\mathbf H^{j,i}}u(x)\right).
\label{eq:prototype}
\end{equation}
Its corresponding classifier input is $h^{\rm ref}_{i,j}=[\mathbf S^{j,i};F_{\theta_i^f}(\mathbf S^{j,i})]$. The target-class row of the cross-entropy classifier-weight gradient is cached after warm-up. This reference captures class-level information for the subsequent correction process.

\noindent
\textBF{The integration optimization process.} Many backdoor data samples would contain perturbations on the class label and image spaces. Correcting these samples with reliable pseudo-labels and integrating them into the training process can enhance the model's robustness against backdoor samples. To achieve this goal, one reasonable approach is to use the learned expert ${\mathcal{W}}_i$ to generate the pseudo-label for a given sample ${\Tilde{\bf x}}$ from the set ${\Tilde{\bf H}}^{j,i}$. However, the presence of adversarial perturbations within the samples precludes any guarantee regarding the correctness of the generated pseudo-labels. To address this issue, the proposed GDBRO mechanism introduces a new gradient discrepancy-based sample selection process that evaluates the similarity between novel instances and class prototypes with identical labels within gradient space. Specifically, for a given sample ${\Tilde{\bf x}} \in {\Tilde{\bf H}}^{j,i}$, the proposed GDBRO mechanism first determines an appropriate expert during the $i$-th task learning by comparing the confidence levels among experts, expressed as~:
\begin{equation}
\begin{aligned}
&F_{\rm find}(\tilde{\bf x})\\
&=\underset{s=1,\ldots,i}{\operatorname{argmax}}
\Big\{\max_{t=1,\ldots,|{\mathcal{Y}}|}
\big\{F_p(\tilde{\bf x},\mathcal W_s)[t]\big\}\Big\}.
\end{aligned}
\end{equation}
\noindent where $s^\star = F_{\rm find}( {\Tilde{\bf x}} )$ denotes the index of the selected expert used for generating the pseudo-label $y' = F_{\rm label}(F_p(\tilde{\bf x},\mathcal W_{s^\star}))$. $F_{\rm label}(\cdot)$ is a function that transfers the probability vector to a class label. In order to assess the class-reference consistency of the pseudo-label $y'$, the proposed GDBRO mechanism evaluates the gradient discrepancy between the given sample ${\Tilde{\bf x}}$ and the corresponding class prototype, expressed as~:
\begin{equation}
\begin{aligned}
F_{\rm disc}(\tilde x,\mathcal W_{s^\star})&=1-F_{\rm cos}(\nabla,\nabla'),\\
\nabla&=\nabla_{W_{s^\star}[y',:]}\mathcal L_{\rm CE}(h_{s^\star}(\tilde x),y'),\\
\nabla'&=g^{\rm ref}_{s^\star,y'}.
\end{aligned}\label{eq:dis}
\end{equation}
\noindent where $y'$ is the predicted class used as the target for both the sample and its cached reference. Only row $y'$ of classifier weights $W_{s^\star}$ is differentiated, rather than the full classifier or model. Samples without an available class-reference gradient are skipped. In the following cross-entropy expression, $\mathbf y$ denotes this one-hot pseudo-label:
\begin{equation}
\begin{aligned}
&{\mathcal{L}}_{\rm CE}({\mathcal{W}}_{s^\star},{\bf z},{\bf y})\\
&\quad=-\sum_{j=1}^{|{\mathcal{Y}}|}{\bf y}[j]\log F_{\theta^c_{s^\star}}({\bf z})[j]\,.
\end{aligned}
\end{equation}
\noindent where $F_{\theta^c_{s^\star}}( {\bf z})[j]$ denotes the $j$-th dimension of the predicted vector returned by $F_{\theta^c_{s^\star}}( {\bf z})$. $F_{\rm cos}(\cdot,\cdot)$ evaluates the cosine similarity between the two target-row gradient vectors, expressed by:
\begin{equation}
\begin{aligned}
  F_{\rm cos}( \nabla,\nabla' ) = \frac{\sum_{j=1}^{|\nabla|} \nabla[j]\nabla'[j]}{\sqrt{\sum_{i=1}^{|\nabla|} ({\nabla}[i])^2} \sqrt{\sum_{i=1}^{|\nabla|} ({\nabla'}[i])^2}}\,,
\end{aligned}
\end{equation}
\noindent where $|\nabla|$ denotes the dimension of the gradient vector $\nabla$. $\nabla[j]$ and $\nabla'[j]$ represent the $j$-th dimension of the gradient vectors $\nabla$ and $\nabla'$, respectively. A high gradient discrepancy score calculated by Eq.~\eqref{eq:dis} indicates that the data ${\Tilde{\bf x}}$ is far away from the semantic feature center of the same class label and suggests weaker support for the generated pseudo-label $y'$. To reduce labeling errors, the proposed GDBRO mechanism forms a robust subset ${\Hat{\bf H}}^i$ by retaining only candidate poisoned samples whose pseudo-labels are supported by low gradient discrepancy~:
\begin{equation}
\begin{aligned}
\hat{\bf H}^i=\big\{&(\tilde{\bf x}_j,\tilde{\bf y}_j)\mid\tilde{\bf x}_j\in\tilde{\bf H}^i,\\
&\tilde{\bf y}_j=F_{\rm label}(F_p(\tilde{\bf x}_j,{\mathcal{W}}_{s^\star_j})),\\
&F_{\rm disc}(\tilde{\bf x}_j,{\mathcal{W}}_{s^\star_j})<\lambda,\\
&s^\star_j=F_{\rm find}(\tilde{\bf x}_j),\;j=1,\ldots,|\tilde{\bf H}^i|\big\}.
\end{aligned}
\end{equation}
\noindent where $\lambda$ is a pre-defined threshold to filter out samples that contain significantly different semantic feature information with respect to the category center. By utilizing the clean and robust datasets ${\bf H}^i$ and ${\Hat{\bf H}}^i$, the proposed GDBRO mechanism integrates them to optimize the current expert ${\mathcal{W}}_i$ during the $i$-th task learning, aiming to improve the model's clean performance and generalization performance~:
\begin{equation}
\mathcal L_{\rm integr}=\mathbb E_{(x,y)\sim P_{\mathbf H^i\cup\hat{\mathbf H}^i}}
[-\log F_p(x,\mathcal W_i)[y]].
\label{eq:joint}
\end{equation}
The joint stage samples the concatenated retained and recovered dataset without a separate recovered-sample loss weight.

\subsection{Robust Feature Consistency-based Expert Selection}

Dynamic-expansion models rely on selecting an appropriate expert at inference time. Under poisoned or perturbed inputs, standard feature-distance-based routing can become unreliable because the sample representation may drift away from clean class prototypes. We address this issue by proposing a novel Robust Feature Consistency-based Expert Selection (RFCBES) that incorporates corrected samples into the class prototypes to improve expert selection.

In order to learn robust representations, once the current task learning is finished (${\mathcal{E}}_i$), the proposed RFCBES mechanism employs the learned expert module ${\mathcal{W}}_i$ to update the sample prototypes by using the joint dataset that contains clean and corrected candidate-poisoned samples. To achieve this goal, we first compute the sample prototype set by~:
\begin{equation}
\begin{aligned}
\tilde{\mathbf S}^{j,i}&=\operatorname{normalize}\left(
\frac{\sum_{x\in\mathbf H^{j,i}\cup\hat{\mathbf H}^{j,i}}u(x)}{|\mathbf H^{j,i}\cup\hat{\mathbf H}^{j,i}|}\right),\\
\tilde{\mathbf S}^i&=\{\tilde{\mathbf S}^{j,i}\}_{j=1}^{c_i}.
\end{aligned}\label{eq:routing_prototypes}
\end{equation}
\noindent where ${\Tilde{\bf S}}^i$ denotes the sample prototype set and is preserved for the expert ${\mathcal{W}}_i$. Each member of ${\Tilde{\bf S}}^i$ captures the statistical information for a unique category. Once the learning process of $n$ tasks is finished, the proposed RFCBES mechanism chooses the best expert for a given testing sample ${\bf x}$ by~:
\begin{equation}
\begin{aligned}
i^\star&=\underset{i=1,\ldots,n}{\operatorname{argmin}}\\
&\quad\Big\{\min_{j=1,\ldots,c_i}\big\{F_{\rm wdis}(\tilde{\bf S}^i[j],u({\bf x}))\big\}\Big\}.
\end{aligned}
\end{equation}
where $\tilde{\mathbf{S}}^i[j]$ denotes the $j$-th instance of the sample prototype set preserved by $\mathcal{W}_i$ and $c_i$ represents the total number of classes within the $i$-th task. $F_{\rm wdis}$ denotes the L2 Euclidean distance between the normalized features. $i^\star$ denotes the index of the selected expert used to make the prediction for the given testing sample $\mathbf{x}$.

The final routing prototypes are recomputed after joint training; the correction-gradient cache retains its warm-up state. These two stores serve different purposes.

\noindent
\textBF{Representation choice and numerical scope.} The BPP and routing features are normalized shared-backbone outputs. The correction feature is the concatenation of shared and expert-specific outputs, while its reference is constructed by passing a normalized class mean through the expert. Comparing a distance in the shared feature space with a class-row gradient distance can therefore compare different feature spaces and different reference normalization. A controlled diagnostic should explicitly identify the two representations and also compare gradient cosine with cosine on the identical classifier-input vectors.

The code uses the cosine-similarity operator with its numerical denominator protection. When a probability saturates or a feature/gradient becomes very small, the exact idealized cancellation need not hold numerically. These cases should be diagnosed separately; they are not evidence of a distinct gradient separation mechanism. No lower noise rate for every selected subset is claimed from the angular criterion alone.

\section{Theoretical Analysis}

This section provides a theoretical analysis of the proposed CLUBA defense
framework. Different from conventional continual learning analysis that mainly
focuses on catastrophic forgetting, our analysis investigates the robustness
of incremental learners against contaminated supervision. Specifically, we
establish theoretical guarantees for three key components: (1) the capability
of Bi-Prototype Purification (BPP) to identify suspicious samples in semantic
feature space, (2) the reliability of Gradient Discrepancy-based Robustness
Optimization (GDBRO) for recovering informative poisoned samples, and (3) the
stability of Robust Feature Consistency-based Expert Selection (RFCBES).

The analysis is based on three mild assumptions regarding feature distribution,
gradient concentration, and prototype separability. These assumptions are
commonly adopted in representation learning and prototype-based classification
analysis.

\subsection{Assumptions and Preliminaries}

Let $u(x)$ denote the normalized shared representation extracted by the frozen
pretrained backbone:

\begin{equation}
u(x)=\frac{F_{\xi}(x)}
{\|F_{\xi}(x)\|_2}.
\end{equation}

For each class $j$ in task $i$, let $\mathcal P_{c}^{j,i}$ and
$\mathcal P_{p}^{j,i}$ denote the distributions of clean and poisoned samples,
respectively.

\noindent
\textbf{Assumption 1 (Semantic Feature Concentration).}
For each class $j$ of task $i$, the shared representations of clean samples
are concentrated around an unknown semantic center $\mu_c^{j,i}$:
\begin{equation}
\mathbb E_{x\sim \mathcal P_c^{j,i}}
\left[
\|u(x)-\mu_c^{j,i}\|_2^2
\right]
\leq\sigma_c^2 .
\end{equation}
Similarly, poisoned samples follow another distribution with center
$\mu_p^{j,i}$ satisfying
\begin{equation}
\mathbb E_{x\sim \mathcal P_p^{j,i}}
\left[
\|u(x)-\mu_p^{j,i}\|_2^2
\right]
\leq\sigma_p^2 .
\end{equation}
The assumption indicates that although backdoor perturbations may be
imperceptible in the original input space, they introduce a semantic deviation
in the learned representation space. Such an assumption is reasonable because
backdoor samples are expected to form distinguishable latent patterns during
training.

\noindent
\textbf{Assumption 2 (Backdoor Semantic Separation).} For each contaminated class, the distance between clean and poisoned semantic
centers satisfies
\begin{equation}
\Delta^{j,i}
=
\|
\mu_c^{j,i}-\mu_p^{j,i}
\|_2
>
\sigma_c+\sigma_p .
\label{eq:semantic_margin}
\end{equation}

\noindent
\textbf{Assumption 3 (Gradient Concentration).} For a correctly pseudo-labeled sample $(x,y)$, the classifier-row gradient
\begin{equation}
g(x,y)
=
\nabla_{W_y}
\mathcal L_{\rm CE}
(h_i(x),y)
\end{equation}
is concentrated around its class reference gradient $g_y^{*}
=
\mathbb E[g(x,y)]$. Specifically, $\|g(x,y)-g_y^{*}\|_2
\leq\epsilon$.

\noindent
\textbf{Definition 1 (Prototype Margin).} For a class prototype set
\begin{equation}
\mathcal S^i
=
\{S^{1,i},...,S^{c_i,i}\},
\end{equation}
the prototype margin of task $i$ is defined as
\begin{equation}
\delta_i
=
\min_{a\neq b}
\|
S^{a,i}-S^{b,i}
\|_2 .
\label{eq:prototype_margin}
\end{equation}
A larger $\delta_i$ indicates better class-level separability.

\subsection{Analysis of Bi-Prototype Purification}

The BPP module performs two-center clustering within each semantic class.
The following theorem shows that when poisoned samples induce a sufficiently
different semantic center, the clustering objective can recover the two
populations with bounded error.

\noindent
\textbf{Lemma 1.} Given a class subset $\mathbf C_i^j$ containing $m_c$ clean samples and
$m_p$ poisoned samples, the empirical centers obtained by two-center clustering
satisfy
\begin{equation}
\|\hat{\mu}_c-\mu_c\|_2
\leq
\sigma_c
\sqrt{
\frac{2\log(2/\rho)}{m_c}
},
\end{equation}
and
\begin{equation}
\|\hat{\mu}_p-\mu_p\|_2
\leq
\sigma_p
\sqrt{
\frac{2\log(2/\rho)}{m_p}
},
\end{equation}
with probability at least $1-\rho$.

\begin{proof} The result follows from the concentration inequality of bounded second-order
moments. Since normalized representations satisfy
$\|u(x)\|_2=1$, the empirical mean is a sub-Gaussian estimator of the true
distribution center. Applying Hoeffding's inequality independently to clean and poisoned subsets
gives
\begin{equation}
P(
\|\hat{\mu}_c-\mu_c\|_2>\eta
)
\leq
2e^{-m_c\eta^2/(2\sigma_c^2)} .
\end{equation}
Solving the above inequality for $\eta$ yields the desired bound. The poisoned
case follows identically.
\end{proof}

\noindent
\textbf{Theorem 1 (BPP Purification Guarantee).} Under Assumptions 1--2, if
\begin{equation}
\Delta^{j,i}
>
2
\left(
\sigma_c
\sqrt{
\frac{2\log(2/\rho)}{m_c}
}
+
\sigma_p
\sqrt{
\frac{2\log(2/\rho)}{m_p}
}
\right),
\label{eq:bpp_condition}
\end{equation}
then the two clusters generated by BPP correspond to the clean and poisoned
populations with probability at least $1-\rho$.

\begin{proof}
According to Lemma 1, the estimated centers satisfy
\begin{equation}
\|\hat{\mu}_c-\mu_c\|
\leq\epsilon_c,
\end{equation}
and
\begin{equation}
\|\hat{\mu}_p-\mu_p\|
\leq\epsilon_p .
\end{equation}
Using the triangle inequality, we have:
\begin{equation}
    \begin{aligned}
\|\hat{\mu}_c-\hat{\mu}_p\|
&\geq
\|\mu_c-\mu_p\|
-\|\hat{\mu}_c-\mu_c\|
-\|\hat{\mu}_p-\mu_p\|
\\
&\geq
\Delta^{j,i}
-\epsilon_c-\epsilon_p .
    \end{aligned}
\end{equation}
According to Eq.~\eqref{eq:bpp_condition}, we have:
\begin{equation}
\|\hat{\mu}_c-\hat{\mu}_p\|
>
\epsilon_c+\epsilon_p .
\end{equation}
Therefore, the empirical centers remain separated after estimation error. Consequently, the optimal two-center clustering assigns samples according to
their generating distributions, completing the proof.
\end{proof}

\noindent
\textbf{Theoretical implication.}
Theorem 1 demonstrates that the effectiveness of BPP is guaranteed by
semantic separability rather than explicit access to poisoning patterns.
When the semantic deviation introduced by contaminated supervision is larger
than the estimation uncertainty of feature prototypes, BPP can preserve the
intrinsic distribution structure and obtain reliable clean--poisoned separation.
Therefore, BPP provides a theoretically justified mechanism for identifying
suspicious samples in the shared representation space.

\subsection{Theoretical Analysis of Gradient Discrepancy-based Robustness Optimization}

After BPP purification, the suspicious subset may still contain informative
samples that are incorrectly separated due to limited feature discrimination.
GDBRO further evaluates these samples from the optimization perspective.
Instead of relying solely on representation similarity, GDBRO measures whether
the gradient induced by a suspicious sample is consistent with the class-level
optimization direction obtained from reliable samples.

The following analysis demonstrates that the proposed gradient discrepancy
criterion provides a theoretically justified reliability measurement for
pseudo-label recovery.

\noindent
\textBF{Definition 2 (Class Gradient Prototype).} For class $j$ of task $i$, the gradient prototype is defined as
\begin{equation}
g^{\mathrm{ref}}_{i,j}
=
\nabla_{W_i[j,:]}
\mathcal{L}_{CE}
(h_i(S^{j,i}),j),
\label{eq:gradient_prototype}
\end{equation}
where $S^{j,i}$ is the semantic class prototype obtained from Eq.~\eqref{eq:prototype}.

\noindent
\textBF{Definition 3 (Gradient Discrepancy).} For a sample $\tilde{x}$ with pseudo-label $\tilde{y}$, the gradient discrepancy
used by GDBRO is defined as
\begin{equation}
D_g(\tilde{x},\tilde{y})
=
1-
\frac{
g(\tilde{x},\tilde{y})^\top
g^{ref}_{i,\tilde{y}}
}
{
\|g(\tilde{x},\tilde{y})\|_2
\|g^{ref}_{i,\tilde{y}}\|_2
}.
\label{eq:def_gradient_discrepancy}
\end{equation}

\noindent
\textbf{Lemma 2.} Under Assumption 3, for a correctly pseudo-labeled clean sample, the gradient
discrepancy satisfies
\begin{equation}
D_g(x,y)
\leq
\frac{
2\epsilon^2
}
{
\|g_y^{*}\|_2^2
},
\label{eq:clean_gradient_bound}
\end{equation}
with probability at least $1-\rho$.
\begin{proof}
Let $g(x,y)=g_y^{*}+\Delta g$
where $\|\Delta g\|_2\leq\epsilon$. The cosine similarity can be rewritten as
\begin{equation}
\cos(g,g_y^{*})
=
\frac{
(g_y^{*}+\Delta g)^\top g_y^{*}
}
{
\|g_y^{*}+\Delta g\|_2
\|g_y^{*}\|_2
}.
\end{equation}
Since $\Delta g$ is bounded, applying the perturbation inequality of cosine
similarity gives
\begin{equation}
1-\cos(g,g_y^{*})
\leq
\frac{2\|\Delta g\|_2^2}
{\|g_y^{*}\|_2^2}.
\end{equation}
Substituting $\|\Delta g\|_2\leq\epsilon$ yields
\begin{equation}
D_g(x,y)
\leq
\frac{2\epsilon^2}
{\|g_y^{*}\|_2^2}.
\end{equation}
\end{proof}

\noindent
\textbf{Lemma 3.} Assume that a poisoned sample generates a gradient deviation satisfying
\begin{equation}
\|
g_p-g_y^{*}
\|_2
\geq
\gamma .
\label{eq:poison_gradient_deviation}
\end{equation}
Then the corresponding gradient discrepancy satisfies
\begin{equation}
D_g(x_p,y)
\geq
\frac{
\gamma^2
}
{
2
\|g_y^{*}\|_2^2
}.
\label{eq:poison_gradient_bound}
\end{equation}

\begin{proof}
Let $g_p=g_y^{*}+\Delta_p$ where $\|\Delta_p\|_2\geq\gamma$. Using the reverse perturbation property of cosine similarity,
\begin{equation}
1-\cos(g_p,g_y^{*})
\geq
\frac{
\|\Delta_p\|_2^2
}
{
2\|g_y^{*}\|_2^2}.
\end{equation}

Since $\|\Delta_p\|_2\geq\gamma$, we obtain
\begin{equation}
D_g(x_p,y)
\geq
\frac{
\gamma^2
}
{
2\|g_y^{*}\|_2^2}.
\end{equation}
\end{proof}

\begin{table*}[t]
    \centering
    \caption{Comprehensive performance comparison of different defenses against various attacks across three datasets. We report the baseline Clean Accuracy (CA, \%) trained on pure clean data, alongside the CA (\%) and Attack Success Rate (ASR, \%) under poisoned scenarios. Each entry reports a single-seed experiment. The best results are highlighted in \textbf{bold}.}
    \vspace{-5pt}
    \label{tab:exp_combined}
    \renewcommand{\arraystretch}{1.2}
    \setlength{\tabcolsep}{3.5pt}
    \resizebox{\textwidth}{!}{
        \begin{tabular}{l ccccccc ccccccc ccccccc}
            \toprule
            \multirow{3}{*}{\textbf{Method}} & \multicolumn{7}{c}{\textbf{CIFAR-10}} & \multicolumn{7}{c}{\textbf{CIFAR-100}} & \multicolumn{7}{c}{\textbf{Tiny-ImageNet}} \\
            \cmidrule(lr){2-8} \cmidrule(lr){9-15} \cmidrule(lr){16-22}
            
            & Clean & \multicolumn{2}{c}{BadNets} & \multicolumn{2}{c}{LTB} & \multicolumn{2}{c}{CBACL} 
            & Clean & \multicolumn{2}{c}{BadNets} & \multicolumn{2}{c}{LTB} & \multicolumn{2}{c}{CBACL} 
            & Clean & \multicolumn{2}{c}{BadNets} & \multicolumn{2}{c}{LTB} & \multicolumn{2}{c}{CBACL} \\
            \cmidrule(lr){2-2} \cmidrule(lr){3-4} \cmidrule(lr){5-6} \cmidrule(lr){7-8} 
            \cmidrule(lr){9-9} \cmidrule(lr){10-11} \cmidrule(lr){12-13} \cmidrule(lr){14-15}
            \cmidrule(lr){16-16} \cmidrule(lr){17-18} \cmidrule(lr){19-20} \cmidrule(lr){21-22}
            
            & CA$\uparrow$ & CA$\uparrow$ & ASR$\downarrow$ & CA$\uparrow$ & ASR$\downarrow$ & CA$\uparrow$ & ASR$\downarrow$ 
            & CA$\uparrow$ & CA$\uparrow$ & ASR$\downarrow$ & CA$\uparrow$ & ASR$\downarrow$ & CA$\uparrow$ & ASR$\downarrow$ 
            & CA$\uparrow$ & CA$\uparrow$ & ASR$\downarrow$ & CA$\uparrow$ & ASR$\downarrow$ & CA$\uparrow$ & ASR$\downarrow$ \\
            \midrule
            
            DER     & 95.86 & 93.79 & 13.07 & 93.94 & 9.17  & 68.11 & 33.83 & 42.97 & 31.31 & 45.08 & 43.03 & 30.99 & 43.58 & 46.12 & 68.16 & 66.74 & 28.19 & 64.20 & 35.14 & 46.80 & 51.18 \\
            DER++   & 93.75 & 94.66 & 33.59 & 84.94 & 15.56 & 90.54 & 34.21 & 39.81 & 33.03 & 40.66 & 61.78 & 42.12 & 35.68 & 41.24 & 70.98 & 72.51 & 22.11 & 68.33 & 29.80 & 54.17 & 43.91 \\
            ER-ACE  & 95.48 & 93.05 & 52.10 & 92.50 & 25.40 & 94.32 & 13.34 & 58.99 & 58.77 & 0.84  & 57.12 & 11.20 & 60.09 & 2.04  & 65.00 & 64.72 & 1.01  & 36.78 & 68.10 & 64.77 & 1.12 \\
            GEM     & 95.89 & 93.73 & 47.22 & 94.10 & 18.20 & 95.67 & 10.80 & 75.85 & 75.29 & 2.45  & 73.50 & 14.50 & 75.27 & 1.88  & 82.42 & 81.25 & 2.93  & 67.57 & 20.25 & 81.27 & 3.60 \\
            A-GEM    & 86.54 & 87.58 & 14.23 & 85.30 & 20.10 & 87.29 & 14.32 & 46.94 & 31.64 & 44.91 & 42.10 & 38.60 & 32.00 & 43.43 & 68.65 & 49.26 & 46.52 & 48.10 & 40.20 & 47.47 & 46.59 \\
            AFLC    & 81.82 & 86.18 & 21.68 & 77.25 & 79.03 & 65.31 & 46.95 & 30.29 & 47.72 & 36.32 & 67.92 & 41.46 & 32.78 & 41.24 & 60.27 & 62.58 & 33.55 & 58.40 & 39.10 & 54.17 & 43.91 \\
            AIR     & 65.57 & 45.30 & 58.58 & 67.22 & 65.31 & 67.81 & 16.76 & 70.08 & 67.27 & 68.80 & 44.60 & 70.69 & 15.56 & 51.85 & 54.41 & 54.75 & 79.71 & 51.20 & 75.40 & 21.32 & 73.28 \\
            FT      & 72.06 & 80.32 & 26.75 & 77.22 & 18.82 & 79.09 & 40.77 & 39.19 & 74.44 & 33.91 & 77.13 & 1.46  & 74.21 & 36.06 & 75.55 & 77.72 & 17.88 & 74.10 & 22.30 & 78.32 & 12.70 \\
            SAM-FT   & 69.08 & 71.09 & 9.70  & 40.50 & 67.53 & 70.62 & 24.41 & 39.19 & 40.68 & 11.17 & 38.50 & 15.40 & 37.10 & 18.20 & 66.77 & 67.75 & 9.48  & 65.20 & 12.50 & 64.10 & 14.20 \\
            AIBD    & 71.72 & 58.49 & 44.19 & 87.00 & 9.43  & 59.59 & 44.26 & 46.19 & 27.30 & 54.70 & 44.20 & 50.10 & 28.08 & 52.36 & 71.77 & 42.27 & 55.12 & 40.10 & 52.30 & 39.87 & 58.14 \\
            RGAN    & 93.96 & 79.87 & 15.79 & 70.44 & 65.11 & 65.29 & 54.77 & 33.76 & 31.94 & 69.62 & 65.29 & 54.77 & 8.78  & 41.51 & 68.05 & 25.08 & 62.90 & 24.10 & 60.15 & 25.02 & 62.93 \\
            \midrule
            \textbf{OURS} & \textbf{99.30} & \textbf{99.10} & \textbf{0.37} & \textbf{99.02} & \textbf{0.31} & \textbf{99.17} & \textbf{0.51} & \textbf{93.20} & \textbf{90.94} & \textbf{0.07} & \textbf{92.15} & \textbf{0.12} & \textbf{91.50} & \textbf{0.45} & \textbf{93.45} & \textbf{91.19} & \textbf{0.15} & \textbf{92.05} & \textbf{1.20} & \textbf{91.80} & \textbf{0.50} \\
            \bottomrule
        \end{tabular}
    }
    \renewcommand{\arraystretch}{1.0}
    
\end{table*}

\noindent
\textbf{Theorem 2 (Gradient-based Recovery Guarantee).} Under Assumptions 1--3, if the gradient deviation satisfies $\gamma>2\epsilon$, there exists a threshold $\lambda$ such that
\begin{equation}
\frac{
2\epsilon^2
}
{
\|g_y^{*}\|_2^2
}
<
\lambda
<
\frac{
\gamma^2
}
{
2\|g_y^{*}\|_2^2
},
\end{equation}
and GDBRO correctly retains clean-supported suspicious samples while rejecting
samples whose pseudo-labels are inconsistent with the class gradient direction.

\begin{proof}
According to Lemma 2, clean-supported samples satisfy
\begin{equation}
D_g(x,y)
\leq
\frac{2\epsilon^2}
{\|g_y^{*}\|_2^2}.
\end{equation}
According to Lemma 3, unreliable poisoned samples satisfy
\begin{equation}
D_g(x_p,y)
\geq
\frac{\gamma^2}
{2\|g_y^{*}\|_2^2}.
\end{equation}
The condition $\gamma>2\epsilon$
guarantees
\begin{equation}
\frac{2\epsilon^2}
{\|g_y^{*}\|_2^2}
<
\frac{\gamma^2}
{2\|g_y^{*}\|_2^2}.
\end{equation}
Therefore a non-empty interval exists for selecting $\lambda$.
Any threshold within this interval separates the two gradient distributions.
\end{proof}

\noindent
\textbf{Theoretical implication.}
Theorem 2 reveals that gradient discrepancy provides an optimization-level
reliability criterion for suspicious samples. Specifically, samples whose
pseudo-labels are consistent with the class optimization direction produce
bounded gradient deviation, while samples inducing conflicting optimization
directions can be separated by an appropriate discrepancy threshold.
Consequently, GDBRO is able to recover informative samples from the suspicious
set without relying solely on semantic similarity.

\subsection{Theoretical Analysis of Robust Feature Consistency-based Expert Selection}

In dynamic expansion-based continual learning, each expert module is optimized
for a specific task. Therefore, selecting an appropriate expert is critical for
maintaining prediction accuracy under task-incremental inference. However,
backdoor perturbations may cause feature deviation and result in unreliable
routing decisions.

RFCBES addresses this problem by constructing robust prototypes using both
purified samples and gradient-verified recovered samples. This section
theoretically analyzes the stability of the proposed routing mechanism.

\noindent
\textBF{Definition 4 (Expert Prototype).} For expert $\mathcal W_i$, the robust prototype of class $j$ is defined as
\begin{equation}
\tilde S^{j,i}
=
\operatorname{normalize}
\left(
\frac{1}
{|\mathbf H^{j,i}\cup\hat{\mathbf H}^{j,i}|}
\sum_{x\in
\mathbf H^{j,i}\cup\hat{\mathbf H}^{j,i}}
u(x)
\right).
\label{eq:expert_prototype_def}
\end{equation}
The corresponding ideal prototype obtained from clean samples is denoted as
$S^{j,i}$.

\noindent
\textbf{Lemma 4.} Assume that the recovered subset satisfies
\begin{equation}
P(y_{\hat H}=y)\geq1-\eta .
\end{equation}
Then the robust prototype error satisfies
\begin{equation}
\|
\tilde S^{j,i}-S^{j,i}
\|_2
\leq
\eta
+
O
\left(
\sqrt{\frac{1}{m}}
\right),
\label{eq:prototype_error}
\end{equation}
\noindent
where $m$ denotes the number of samples used for prototype estimation.

\begin{proof}

Let $\tilde S^{j,i}
=
(1-\eta)\bar S_c
+
\eta \bar S_r$, where $\bar S_c$ represents the mean feature of correctly recovered samples
and $\bar S_r$ represents the contribution from incorrectly recovered samples.

The deviation from the clean prototype satisfies
\begin{equation}
    \begin{aligned}
\|
\tilde S^{j,i}-S^{j,i}
\|
&=
\|
(1-\eta)\bar S_c+\eta\bar S_r-S^{j,i}
\|
\\
&\leq
(1-\eta)
\|\bar S_c-S^{j,i}\|
+
\eta
\|\bar S_r-S^{j,i}\|. 
    \end{aligned}
\end{equation}
Since normalized features satisfy $\|\bar S_r-S^{j,i}\|\leq2$, we obtain
\begin{equation}
\|
\tilde S^{j,i}-S^{j,i}
\|
\leq
O
\left(
\sqrt{\frac1m}
\right)
+
2\eta .
\end{equation}
Ignoring constant scaling gives Eq.~\eqref{eq:prototype_error}.
\end{proof}

\noindent
\textbf{Theorem 3 (RFCBES Routing Stability).} Assume that the ideal expert prototypes satisfy a minimum inter-expert margin
\begin{equation}
\Delta_r
=
\min_{i\neq k}
\|
S_i-S_k
\|_2 .
\label{eq:routing_margin}
\end{equation}
If the prototype estimation error satisfies
\begin{equation}
\eta+
O(\sqrt{\frac1m})
<
\frac{\Delta_r}{2},
\label{eq:routing_condition}
\end{equation}
then RFCBES selects the correct expert for a test sample whose feature noise
is bounded by
\begin{equation}
\|u(x)-S_i\|_2
<
\frac{\Delta_r}{2}
-
\eta .
\end{equation}

\begin{proof}
For a sample belonging to expert $i$, the routing rule selects
\begin{equation}
i^\star
=
\arg\min_k
\|
u(x)-\tilde S_k
\|_2 .
\end{equation}
For the correct expert, we have
\begin{align}
\|u(x)-\tilde S_i\|
&\leq
\|u(x)-S_i\|
+
\|S_i-\tilde S_i\|
<
\frac{\Delta_r}{2}.
\end{align}
For any incorrect expert $k$, we have
\begin{equation}
    \begin{aligned}
\|u(x)-\tilde S_k\|
&\geq
\|S_i-S_k\|
-
\|u(x)-S_i\|
-
\|S_k-\tilde S_k\|
\\
&>
\Delta_r
-
\frac{\Delta_r}{2}
-
\frac{\Delta_r}{2}
\\
&>
\|u(x)-\tilde S_i\|.
    \end{aligned}
\end{equation}
Therefore, we obtain $i^\star=i.$
\end{proof}

\noindent
\textbf{Theoretical implication.}
Theorem 3 establishes the stability of RFCBES under bounded prototype
estimation errors. The result indicates that robust prototype construction
using purified and gradient-verified samples can effectively preserve the
inter-expert discrimination margin. Therefore, RFCBES maintains consistent
expert routing decisions even when incremental training data contains
contaminated supervision.

\section{Experiments}

\noindent \textbf{Benchmark Datasets and Attack Configurations.} We evaluate our framework on three standard class-incremental benchmarks: \textbf{Split CIFAR-10} (5 tasks), \textbf{Split CIFAR-100} (10 tasks), and \textbf{Split Tiny-ImageNet} (20 tasks). We evaluate against three representative continual backdoor attacks: \textbf{BadNets}~\cite{badnets}, \textbf{LTB}~\cite{persistent_backdoor}, and \textbf{CBACL}~\cite{cbacl}. The backdoor injection rate is fixed at 5\% across all tasks. Detailed descriptions of the datasets and attack mechanisms are provided below.

\noindent \textbf{Comparison Baselines and Implementation Details.} We compare our method against 11 representative baselines, including vanilla continual learning approaches (DER, DER++~\cite{der}, ER-ACE~\cite{er_ace}, GEM~\cite{gem}, A-GEM~\cite{agem}), adversarial defenses (AFLC~\cite{aflc}, AIR~\cite{air}), and backdoor defenses (FT~\cite{ft}, SAM-FT~\cite{samft}, AIBD~\cite{adversarial}, RGAN~\cite{rgan}). All methods utilize a shared \textbf{Pre-trained ViT} backbone. For replay-based methods, the memory buffer size is strictly limited to 500. All models are trained using the SGD optimizer with a learning rate of 0.001 and a batch size of 32, using a requested epoch argument of 1. The proposed method additionally performs five default warm-up epochs, giving five warm-up epochs plus one joint epoch. Training budgets and the repeated-run protocol are specified below.

\noindent \textbf{Evaluation Metrics.} We adopt \textbf{Clean Accuracy (CA)} to measure the retention of benign knowledge (higher is better) and \textbf{Attack Success Rate (ASR)} to measure the efficacy of the backdoor defense (lower is better). Both metrics are evaluated on the test sets of all learned tasks after the entire sequential training process is completed.

\subsection{Datasets and Preprocessing}
We comprehensively evaluate our defense framework on three standard image classification benchmarks tailored for Continual Backdoor Attacks (CBA): Split CIFAR-10, Split CIFAR-100, and Split Tiny-ImageNet. 

In our continual learning configuration, we divide each dataset into $N$ non-overlapping subsets, with each subset corresponding to a sequentially arriving task $\mathcal{T}_i$. The detailed task division and class allocations for each dataset are summarized in Table \ref{tab:dataset_division}.

\begin{table}[t]
    \centering
    \caption{Detailed division of continual learning benchmarks.}
    \vspace{-5pt}
    \label{tab:dataset_division}
    \setlength{\tabcolsep}{1.5pt}
        \begin{tabular}{l ccc}
            \toprule
            \textbf{Dataset} & \textbf{Total Classes} & \textbf{Number of Tasks} & \textbf{Classes per Task} \\
            \midrule
            Split CIFAR-10 & 10 & 5 & 2 \\
            Split CIFAR-100 & 100 & 10 & 10 \\
            Split Tiny-ImageNet & 200 & 20 & 10 \\
            \bottomrule
        \end{tabular}
\end{table}

To accommodate the input dimensional requirements of the shared pre-trained ViT-B/16 backbone, all original images must be standardized. During the training phase, standard data augmentation techniques are applied to prevent overfitting, while deterministic resizing is applied during the testing phase. Table \ref{tab:dataset_transformations} details the original resolutions alongside the specific preprocessing protocols applied across different datasets.

\noindent
\textBF{Data and model attribution.} The experiments use the publicly released CIFAR-10 and CIFAR-100 datasets, Tiny-ImageNet derived from ImageNet, and pre-trained ViT weights. The CIFAR release is available at \url{https://www.cs.toronto.edu/~kriz/cifar.html}. These existing assets provide the benchmarks and shared representation used throughout the evaluation.

\subsection{Framework and Architectural Configurations}
\textbf{Pre-trained ViT Backbone.} Our framework and all baseline methods utilize a unified pre-trained Vision Transformer (\textbf{ViT-B/16}) as the shared backbone. The backbone leverages publicly available weights pre-trained on ImageNet-1K. During the sequential learning process, the shared backbone's parameters are entirely frozen to act as a robust universal feature extractor, mitigating initial vulnerability to poisoned features.

\textbf{Dynamic Expert Modules.} For each newly arriving task $\mathcal{T}_i$, our framework dynamically instantiates a dedicated expert module $W_i$. Each expert comprises a lightweight fully-connected transformation layer and a linear classification head. Upon the completion of task $\mathcal{T}_i$, the parameters of its corresponding expert are strictly frozen. This parameter isolation prevents subsequent gradient updates from modifying the previous experts; inference additionally depends on the routing bank.

\subsection{Backdoor Attack Configurations}
\label{sec:attack_configs}
To rigorously evaluate the robustness of the proposed framework, we implement three state-of-the-art backdoor attacks tailored for or adapted to the continual learning (CL) paradigm. Across all attacks, the poisoning rate is fixed at 5\% for each infected task.

\textbf{BadNets:} We adopt the standard BadNets attack with a static visual trigger. For Split CIFAR-10/100, a $3 \times 3$ white square patch is injected into the bottom-right corner of the images. For Split Tiny-ImageNet, a $4 \times 4$ patch is used. The attack operates under a \textit{label-flip} setting, where the labels of poisoned samples are forcefully altered to a predefined target class .
    
\textbf{Latent Task Backdoor (LTB):} Specifically designed for continual learning, LTB achieves persistence by embedding triggers into the most stable neurons of the model. We utilize a static patch trigger combined with a \textit{label-flip} strategy mapping to a target class. The poisoning strategy explicitly targets the parameter optimization of the current task to ensure the backdoor survives subsequent benign task updates.
    
\textbf{Controllable Backdoor Attack in CL (CBACL):} CBACL leverages an anchor-based representation manipulation strategy. We configure CBACL with a blended trigger (blending a pattern with a 20\% ratio). It operates under a \textit{label-flip} setting targeting a specific class. During training, it utilizes a Triplet Margin Loss to forcefully map poisoned features into a tight target feature space, actively resisting catastrophic forgetting of the backdoor mapping.

\begin{table*}[t]
    \centering
    \caption{Ablation study on the core components of our proposed defense framework. Experiments are performed on CIFAR-100 under the BadNets attack. \cmark~indicates the component is enabled, while \xmark~indicates it is disabled. We report the mean $\pm$ standard deviation over three random seeds (42, 256, and 512).}
    \vspace{-5pt}
    \label{tab:ablation_macro}
        \setlength{\tabcolsep}{13.5pt}
        \begin{tabular}{l ccc cc}
            \toprule
            \multirow{2}{*}{\textbf{Method Variant}} & \multicolumn{3}{c}{\textbf{Components}} & \multicolumn{2}{c}{\textbf{Metrics}} \\
            \cmidrule(lr){2-4} \cmidrule(lr){5-6}
             & \textbf{Dynamic Expert} & \textbf{Feature Clustering} & \textbf{Label Correction} & \textbf{CA (\%)} $\uparrow$ & \textbf{ASR (\%)} $\downarrow$ \\
            \midrule
            Baseline (DER++) & \xmark & \xmark & \xmark & 42.90 $\pm$ 1.32 & 35.80 $\pm$ 3.25 \\
            w/o Dynamic Expert & \xmark & \cmark & \cmark & 34.98 $\pm$ 2.37 & 0.10 $\pm$ 0.03 \\
            w/o Defense Pipeline & \cmark & \xmark & \xmark & 89.87 $\pm$ 0.46 & 27.54 $\pm$ 2.15 \\
            w/o Label Correction & \cmark & \cmark & \xmark & \textbf{91.36} $\pm$ 0.58 & 0.14 $\pm$ 0.09 \\
            \midrule
            \textbf{Ours (Full Model)} & \cmark & \cmark & \cmark & 91.16 $\pm$ 0.33 & \textbf{0.08 $\pm$ 0.10} \\
            \bottomrule
        \end{tabular}    
\end{table*}

\subsection{Training Protocol and Repeated Runs}
Experiments use SGD, a learning rate of 0.001, and minibatches of 32. The requested epoch argument is 1. The proposed method first warms up the current expert for $E_{\rm warm}=5$ epochs and then performs $\max(1,E_{\rm requested}-E_{\rm warm})=1$ joint epoch. Replay-based baselines use a capacity of 500 exemplars. Our method stores experts, class prototypes, and cached class-reference gradients instead of historical training images.

The main comparison is based on one training seed. The macro component ablation reports mean and standard deviation across seeds 42, 256, and 512. K-Means uses two clusters, ten initializations, and a fixed clustering random state of 42; this clustering state is separate from the network-training seeds. Diagnostic studies specify their correction threshold and warm-up settings individually below.

\subsection{Evaluation Metrics}
\label{sec:metrics}
To comprehensively evaluate the robustness of our defense framework and its ability to mitigate catastrophic forgetting under persistent continual backdoor attacks, we employ two primary evaluation metrics: Average Clean Accuracy (CA) and Average Attack Success Rate (ASR). Both metrics are evaluated on the test sets of all learned tasks after the entire sequential training process is completed.

\vspace{5pt}
\noindent
\textbf{Average Clean Accuracy (CA):}~
The Clean Accuracy measures the model's ability to retain benign knowledge and represents the overall classification performance on clean test samples. Specifically, it is defined as the average accuracy on the clean test sets across all $N$ learned tasks:
\begin{equation}
\text{CA} = \frac{1}{N} \sum_{i=1}^{N} A_{i}^{\text{clean}},
\end{equation}
where $N$ is the total number of tasks, and $A_{i}^{\text{clean}}$ is the accuracy of the model on the clean test set of the $i$-th task. A higher CA implies that the model exhibits strong plasticity for learning new sequential tasks while effectively resisting catastrophic forgetting of previously acquired benign knowledge.

\vspace{5pt}
\noindent
\textbf{Average Attack Success Rate (ASR):}~
To strictly evaluate the model's vulnerability against continuous data poisoning, we use the Attack Success Rate. ASR measures the average proportion of poisoned test samples (i.e., inputs embedded with malicious triggers) that are successfully misclassified into the attacker-specified target class across all tasks. It is formulated as:
\begin{equation}
\text{ASR} = \frac{1}{N} \sum_{i=1}^{N} \left( \frac{M_{i}^{\text{poison}}}{T_{i}^{\text{poison}}} \right) \times 100\%,
\end{equation}
where $M_{i}^{\text{poison}}$ denotes the number of poisoned samples successfully misclassified as the target label in task $i$, and $T_{i}^{\text{poison}}$ represents the total number of poisoned test samples in task $i$. A highly robust defense mechanism is characterized by an ASR close to 0\%, indicating that the model successfully neutralizes the backdoor perturbations and refuses to output the malicious target label during inference.

\subsection{Main Results}

Table \ref{tab:exp_combined} summarizes the comprehensive performance comparison of our proposed defense framework against existing CL baselines under three distinct backdoor attacks across CIFAR-10, CIFAR-100, and Tiny-ImageNet datasets.

\textbf{Vulnerability of Baseline Models.} As observed, standard CL methods severely lack robustness when confronted with continual backdoor attacks. For instance, while methods like ER-ACE and GEM maintain relatively high CA on CIFAR-10, their ASR scores soar beyond 40\% under the BadNets attack. In more complex datasets like CIFAR-100 and Tiny-ImageNet, conventional methods experience catastrophic failures. Advanced attacks such as LTB and CBACL consistently deceive the standard baselines, forcing significant drops in CA or pushing the ASR to perilous levels (e.g., AIR exhibits an ASR of 79.71\% under BadNets on Tiny-ImageNet). This indicates that without an explicit purification mechanism, standard dynamic or regularization-based CL frameworks inherently absorb corrupted supervision.

\textbf{Effectiveness of the Proposed Framework.} In contrast, our proposed method demonstrates superior and stable defense capabilities. Across all three datasets and attack paradigms, our model consistently suppresses the ASR to near-zero levels while maintaining high plasticity. Notably, on CIFAR-100 against the BadNets attack, our method reduces the ASR to $0.07\%$ with a CA of $90.94\%$. Furthermore, against state-of-the-art continual backdoor attacks like LTB and CBACL, our framework effectively rejects poisoned perturbations while maintaining high clean accuracy, achieving class-leading CA performance on complex datasets. Moreover, the strictly isolated parameters in our dynamic expert architecture prevent new-task updates from changing earlier expert parameters. Overall task-free predictions also depend on routing through the expanding prototype bank.

\subsection{Ablation Study}

\textbf{Effectiveness of Core Components.} Table \ref{tab:ablation_macro} summarizes the macro-ablation results under BadNets on CIFAR-100 (three-seed mean \ensuremath{\pm} std; seeds 42, 256, and 512). Removing the Dynamic Expert leads to severe forgetting (CA $34.98\% \pm 2.37\%$), while keeping the defense keeps ASR low ($0.10\% \pm 0.03\%$). Omitting the Defense Pipeline raises ASR to $27.54\% \pm 2.15\%$ despite high CA ($89.87\% \pm 0.46\%$). Disabling Label Correction yields the highest CA ($91.36\% \pm 0.58\%$) but a slightly increased ASR ($0.14\% \pm 0.09\%$). Our full model achieves the best trade-off, with CA $91.16\% \pm 0.33\%$ and the lowest ASR $0.08\% \pm 0.10\%$, demonstrating robust, low-variance defense. The DER++ baseline, with CA $42.90\% \pm 1.32\%$ and ASR $35.80\% \pm 3.25\%$, further validates the effectiveness of each proposed component.

\begin{figure*}[t]
    \centering
    \subfloat[Average Forgetting\label{fig:forgetting}]{%
\begin{minipage}[c]{0.32\linewidth}
\centering
        \includegraphics[width=\linewidth]{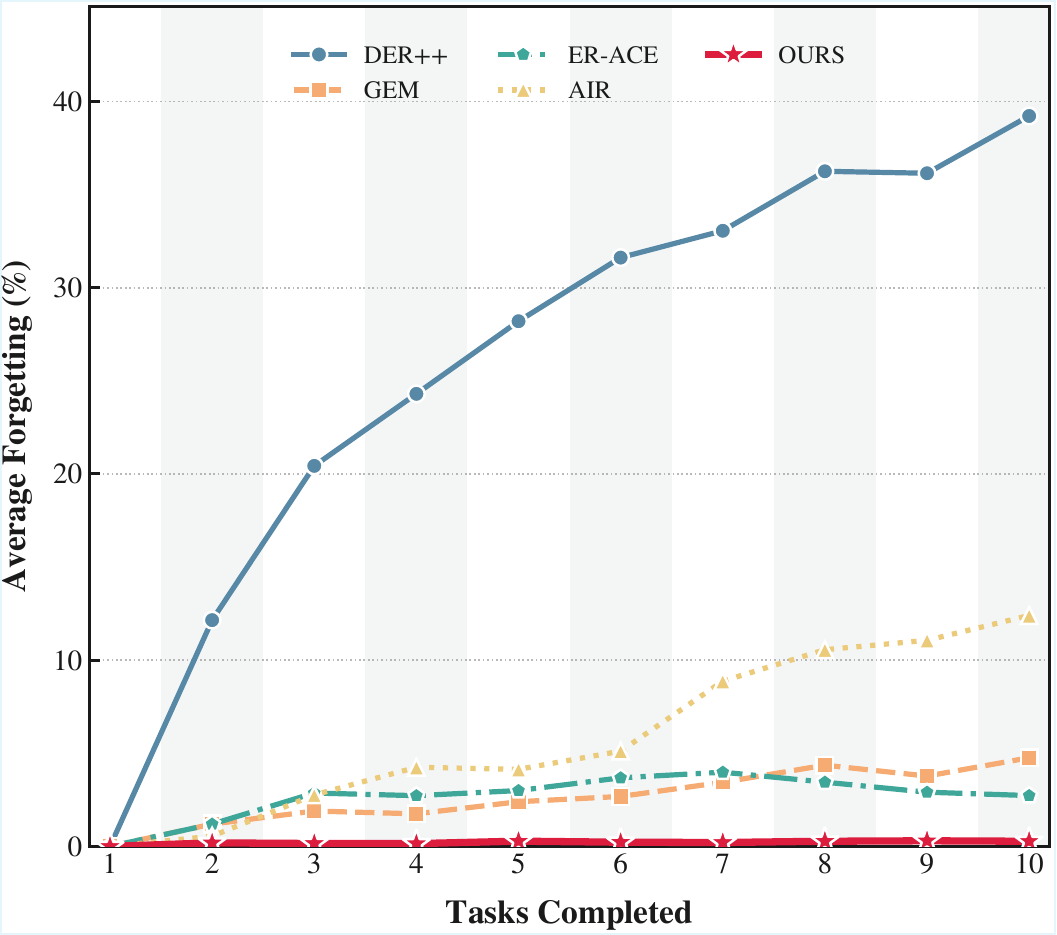}
\end{minipage}}
    \hfill
    \subfloat[Gradient vs. Feature Distance\label{fig:scatter}]{%
\begin{minipage}[c]{0.32\linewidth}
\centering
        \includegraphics[width=0.95\linewidth]{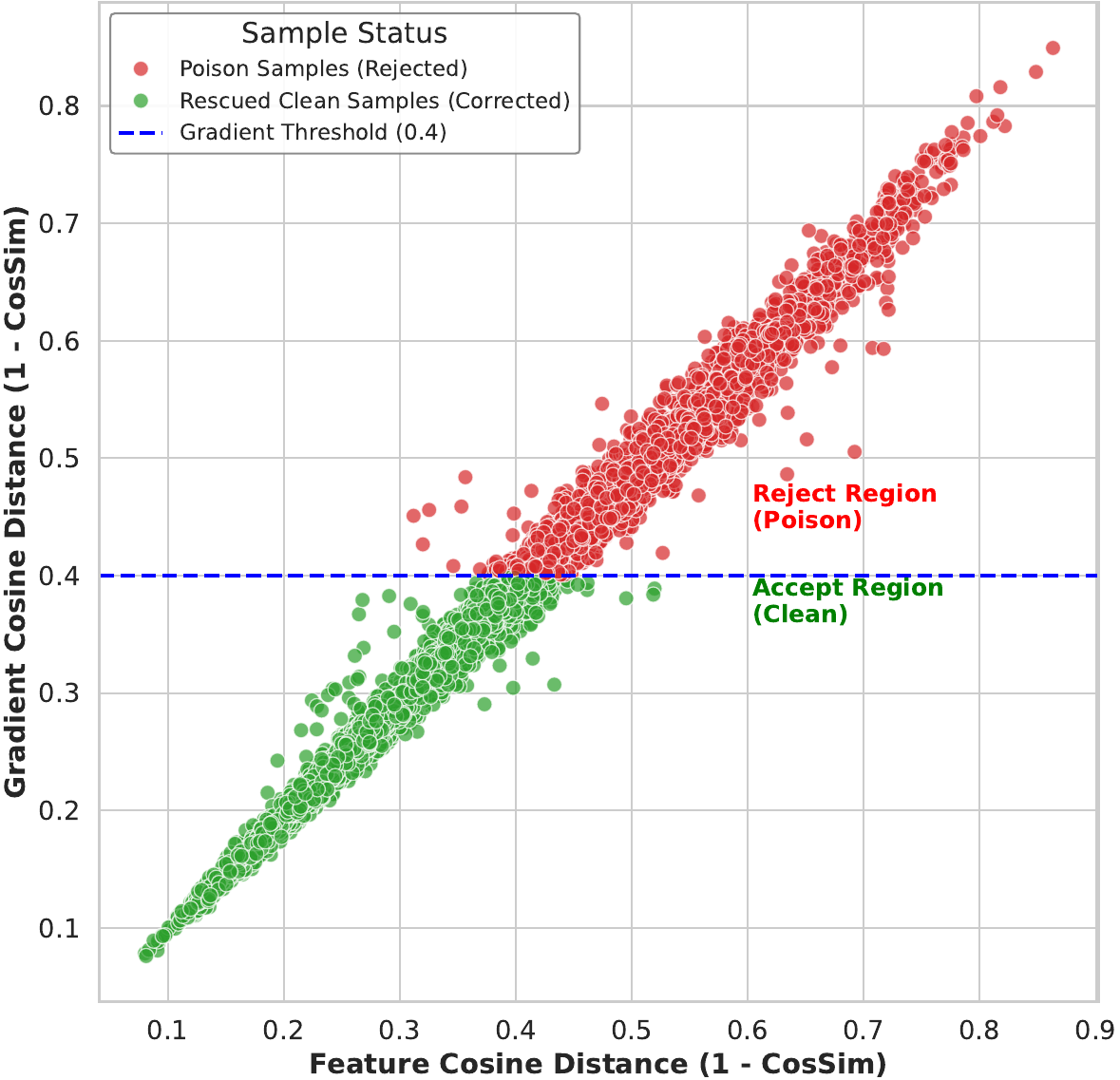}
\end{minipage}}
    \hfill
    \subfloat[Parameters vs. Performance\label{fig:params}]{%
\begin{minipage}[c]{0.32\linewidth}
\centering
        \includegraphics[width=\linewidth]{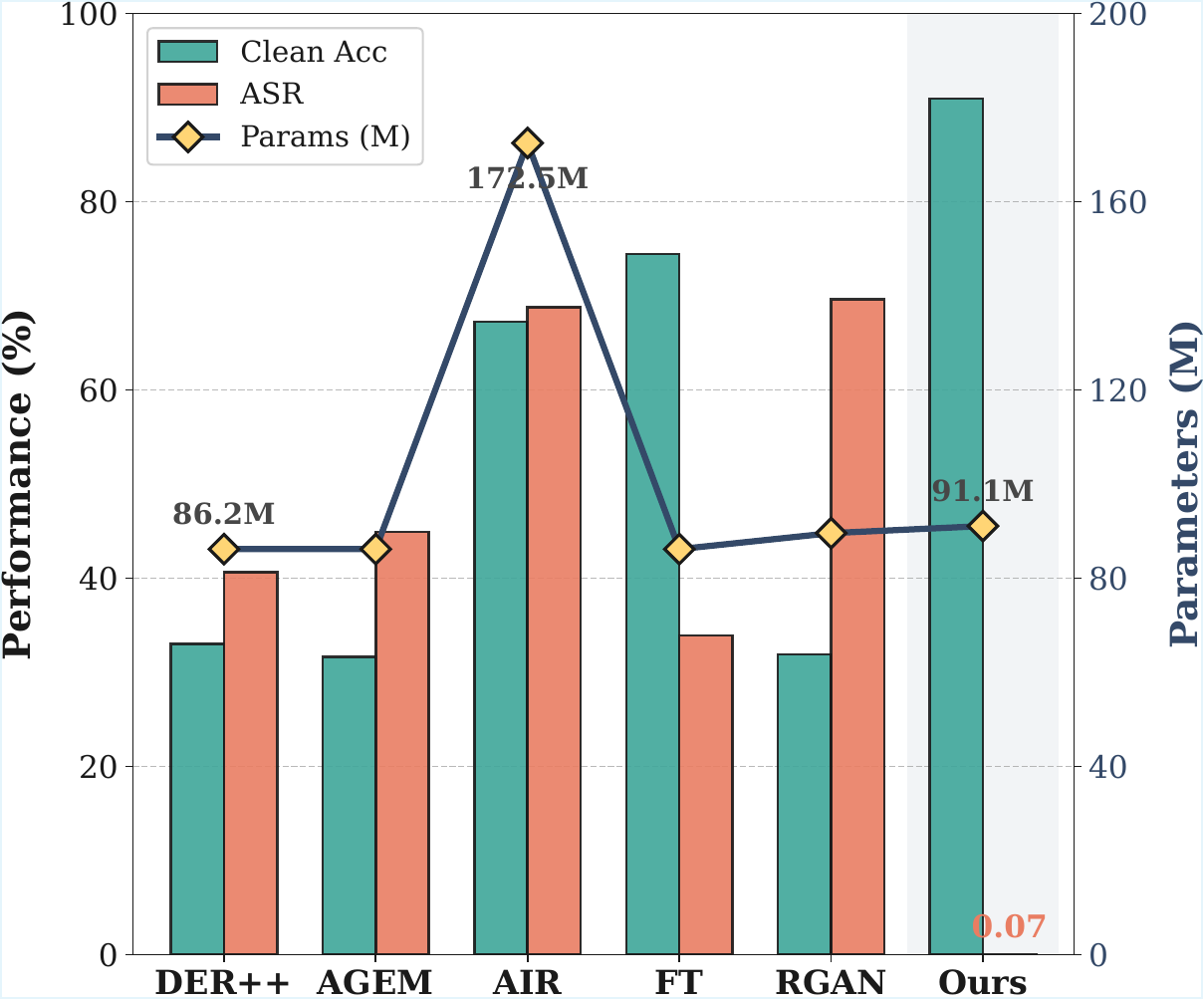}
\end{minipage}}
    
    \caption{Visual analysis of our framework. (a) Average forgetting rate across tasks. (b) Gradient vs. Feature distance distribution during the correction phase. (c) Trade-off between parameter overhead and defense efficacy.}
    \label{fig:visual_analysis}
    \vspace{-5pt}
\end{figure*}

\textbf{Why Gradient-guided Correction?} We conduct this analysis on CIFAR-100 under BadNets and visualize feature distance versus gradient distance of suspicious samples in Figure~\ref{fig:visual_analysis}(b). The reported feature-distance scores overlap, while the gradient-based correction score shows separation at a threshold of 0.4.

\subsection{Analytical Results}

\textbf{Experimental Setup.} The macro and step-wise component ablations are evaluated on the CIFAR-100 dataset with the BadNets attack. We report the average clean accuracy (CA, \%) and attack success rate (ASR, \%) over three seeds for the macro ablation; the main table reports single-seed results.

\noindent
\textbf{Resistance to Catastrophic Forgetting.} Figure \ref{fig:visual_analysis}(a) visualizes the average forgetting rate across sequentially arriving tasks. While standard continual learning methods such as DER++ and AIR suffer from severe catastrophic forgetting (with DER++ exceeding a 40\% forgetting rate by the final task), our framework effectively mitigates this issue. By dynamically expanding lightweight experts tailored to task-specific distributions, our method maintains a near-zero average forgetting rate, preserving previously acquired knowledge remarkably well.

\noindent
\textbf{Parameter Efficiency and Performance Trade-off.} A critical concern in dynamic network architectures is the potential explosion of model parameters. Figure \ref{fig:visual_analysis}(c) highlights the trade-off between parameter overhead and defense efficacy. Methods like AIR drastically inflate the parameter count (up to 172.5M) yet still fail to provide satisfactory robustness. In contrast, our framework leverages a shared Pre-trained ViT backbone combined with highly compact task-specific expert layers, resulting in a minimal parameter increase (91.1M compared to the 86.2M of baseline architectures). Despite this lightweight footprint, our model achieves the highest Clean Accuracy while suppressing the ASR to a negligible $0.07\%$, proving its exceptional parameter efficiency and practical deployment viability.

\subsection{Step-wise Ablation Experiments}

\begin{table}[t]
    \centering
    \caption{Detailed ablation study on the step-wise integration of the proposed mechanisms on CIFAR-100 under BadNets. \textbf{BPP}: Bi-Prototype Purification; \textbf{Warm-up}: Warm-up optimization; \textbf{Relabel}: Gradient-guided label correction; \textbf{RFCBES}: Robust Feature Consistency-based Expert Selection. The best CA and ASR are highlighted in \textbf{bold}.}
    \vspace{-5pt}
    \label{tab:stepwise_ablation}
    \renewcommand{\arraystretch}{1.2}
    \setlength{\tabcolsep}{12.5pt}
{\begin{tabular}{lcc}
        \toprule
        \textbf{Method Variants} & \textbf{CA (\%)} $\uparrow$ & \textbf{ASR (\%)} $\downarrow$ \\
        \midrule
        BPP Only & 55.68 & 3.22 \\
        \quad + Warm-up & 61.78 & 1.19 \\
        \quad + Warm-up + Relabel & \textbf{90.67} & 1.47 \\
        \midrule
        \textbf{\quad + RFCBES (Full Model)} & 90.21 & \textbf{0.07} \\
        \bottomrule
    \end{tabular}}
\end{table}

\textbf{Experimental Setup.} Before delving into the component-wise analysis, we briefly outline the experimental configurations for this ablation study. The experiments are conducted on the Split CIFAR-100 (Seq-CIFAR-100) dataset under the highly challenging Class-Incremental Learning (Class-IL) setting, where task identities are strictly unavailable during inference. We employ the BadNets attack on CIFAR-100. The defense hyperparameters are kept constant across all variants to ensure fairness: the warm-up period is set to 5 epochs, and the gradient correction threshold is fixed at 0.5. To systematically evaluate the contribution of each algorithmic design in our framework, we conduct a step-wise (additive) ablation study, transitioning from a naive baseline to the full model. The results are summarized in Table \ref{tab:stepwise_ablation}.

\begin{figure*}[t]
    \centering
    \includegraphics[width=0.85\textwidth]{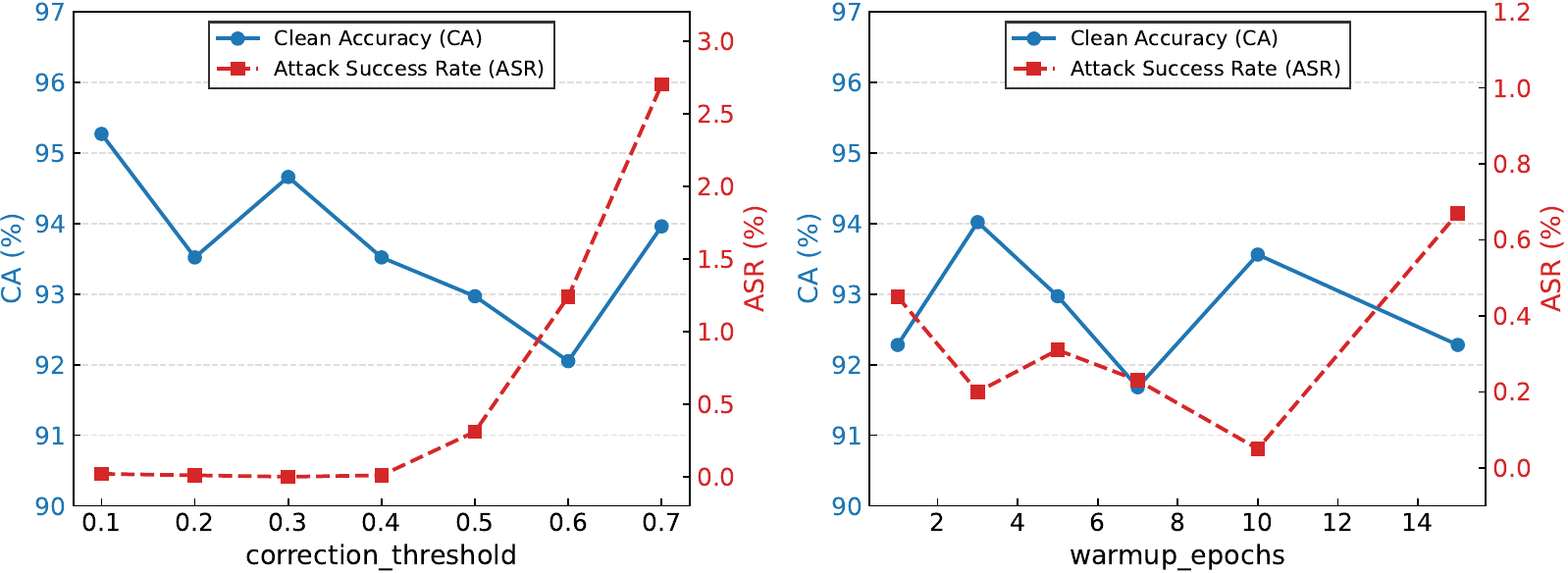}
    \caption{Hyperparameter sensitivity analysis. Left: Effect of gradient discrepancy threshold $\lambda$ (fix $E_{\text{warm}}=5$). Right: Effect of warm-up epochs $E_{\text{warm}}$ (fix $\lambda=0.5$).}
    \label{fig:hyper_sensitivity}
\end{figure*}

\textbf{The baseline purification (BPP Only):} Relying solely on the Bi-Prototype Purification (BPP) mechanism effectively discards the suspicious feature clusters, leading to a preliminary suppression of backdoors (ASR stays below 5\% at 3.22\%). However, this hard-rejection strategy inevitably discards a considerable amount of hard-to-learn but benign samples. Consequently, the model suffers from underfitting on clean distributions, resulting in a suboptimal Clean Accuracy (CA) of 55.68\%.

\noindent
\textbf{The effect of Warm-up optimization:} Introducing the warm-up phase on candidate clean subsets stabilizes the initial feature representations of the newly generated expert module. This optimization significantly boosts the CA by 6.10 percentage points (from 55.68\% to 61.78\%), demonstrating the necessity of establishing a reliable initial optimization trajectory before tackling complex or noisy training instances. The ASR also decreases from 3.22\% to 1.19\%, showing improved resistance in this comparison.

\noindent
\textbf{The power of Gradient-guided Relabeling:} By applying gradient-guided label correction to rescue misclassified clean samples from the rejected subsets, the model reaches its peak CA of 90.67\%. This supports our hypothesis that gradient discrepancy effectively distinguishes and recovers benign instances, enriching the training set and supporting plasticity during continual learning. Nevertheless, this enhanced plasticity comes with a minor security trade-off, slightly raising the ASR to 1.47\% consistent with a less conservative recovery decision.

\noindent
\textbf{The necessity of RFCBES (Full Model):} To perfectly address the aforementioned security trade-off, our Robust Feature Consistency-based Expert Selection (RFCBES) acts as the ultimate gatekeeper during the testing phase. Backdoor attacks typically rely on generating abnormally high softmax confidences to hijack the routing mechanism in Dynamic-expansion Models. By switching from confidence-based routing to prototype-distance-based routing (RFCBES), backdoor triggers completely fail to deceive the expert selection. Consequently, the ASR drops drastically from 1.47\% to a near-zero \textbf{0.07\%}, while maintaining an excellent CA of 90.21\%. 

In conclusion, the training-stage components (Warm-up and Relabeling) predominantly improve the model's plasticity and CA, whereas the testing-stage routing mechanism (RFCBES) reduces the remaining ASR in this experiment.

This step-wise study is reported separately from the three-seed macro ablation; their absolute values are not pooled.

\begin{figure*}[t]
    \centering
    \renewcommand{\thesubfigure}{\Alph{subfigure}}
    \subfloat[DER++\label{fig:overhead_selected_derpp}]{%
\begin{minipage}[b]{\textwidth}
\centering
        \includegraphics[width=\textwidth,height=0.2\textheight,keepaspectratio]{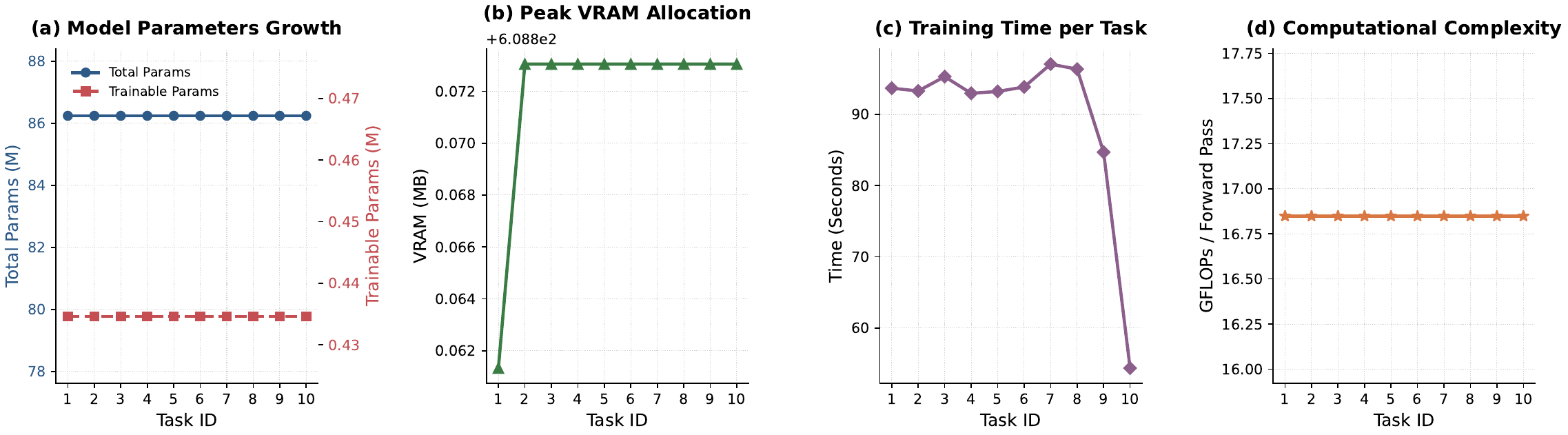}
\end{minipage}}
    \par\vspace{0.1cm}
    \subfloat[AIBD\label{fig:overhead_selected_aibd}]{%
\begin{minipage}[b]{\textwidth}
\centering
        \includegraphics[width=\textwidth,height=0.2\textheight,keepaspectratio]{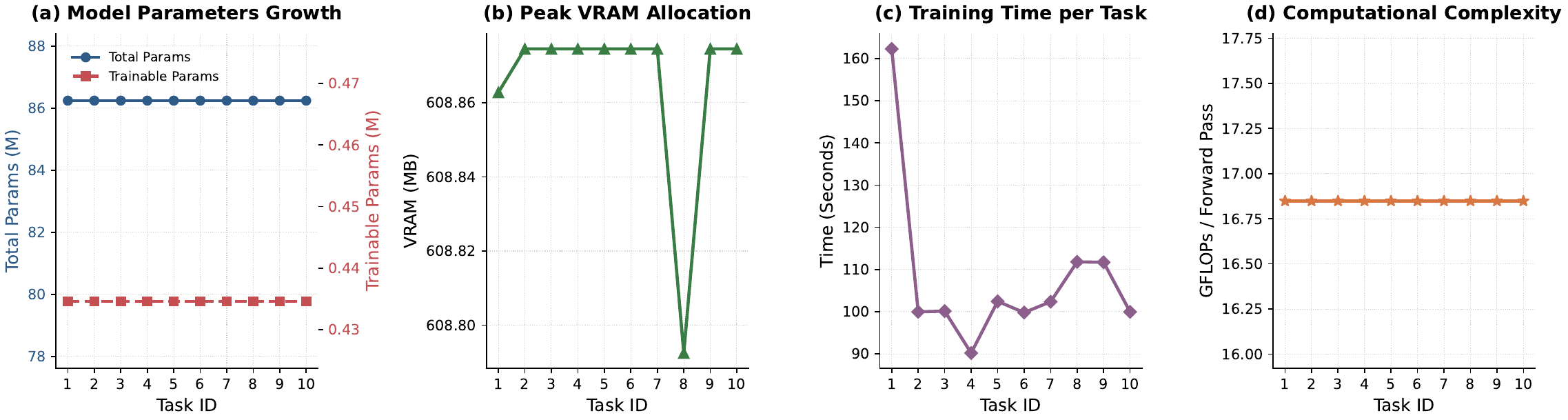}
\end{minipage}}
    \par\vspace{0.1cm}
    \subfloat[Ours (BPP-GDBRO)\label{fig:overhead_selected_ours}]{%
\begin{minipage}[b]{\textwidth}
\centering
        \includegraphics[width=\textwidth,height=0.2\textheight,keepaspectratio]{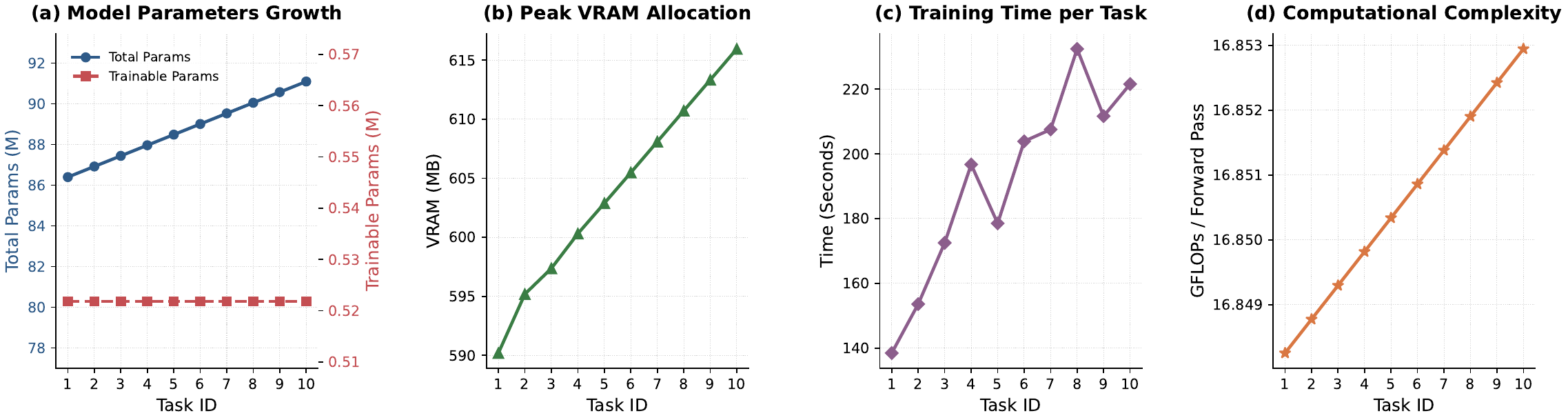}
\end{minipage}}
    \caption{Computational overhead on Split CIFAR-100 for \textbf{(A)} DER++, \textbf{(B)} AIBD, and \textbf{(C)} our BPP-GDBRO framework. Each original plot reports \textbf{(a)} total and trainable parameters, \textbf{(b)} peak GPU memory, \textbf{(c)} training time per task, and \textbf{(d)} GFLOPs per forward pass.}
    \label{fig:overhead_selected}
\end{figure*}

\subsection{Computational Overhead Analysis}

We evaluate practical deployment cost on Split CIFAR-100 by profiling total and trainable parameters, peak GPU memory, training time per task, and GFLOPs per forward pass. The reported setup uses a pre-trained ViT backbone, an NVIDIA RTX 3090 GPU, and a batch size of 32. Figure~\ref{fig:overhead_selected} presents our method alongside two representative baselines: DER++ for rehearsal-based continual learning and AIBD for backdoor defense.

\begin{figure*}[t]
    \centering
    \subfloat{%
\begin{minipage}[c]{0.31\textwidth}
\centering
        \includegraphics[width=\textwidth]{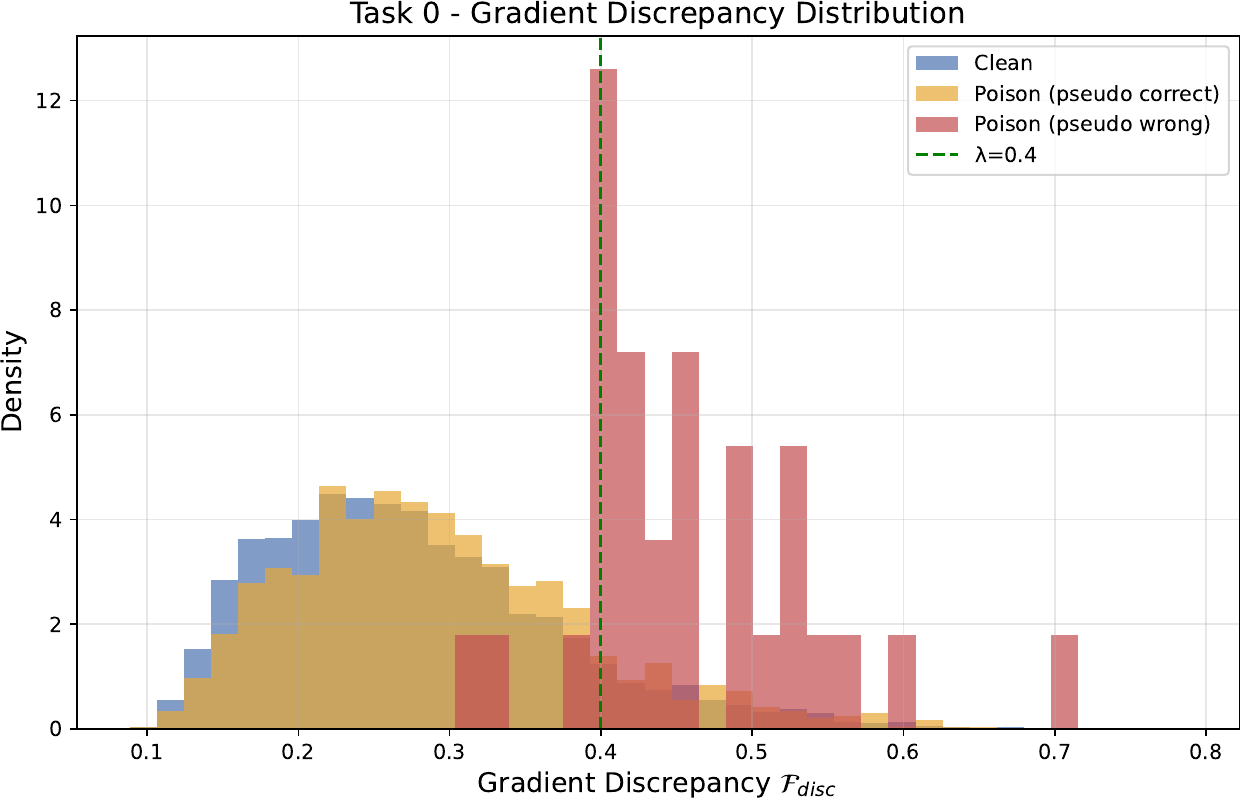}
\end{minipage}}
    \hfill
    \subfloat{%
\begin{minipage}[c]{0.31\textwidth}
\centering
        \includegraphics[width=\textwidth]{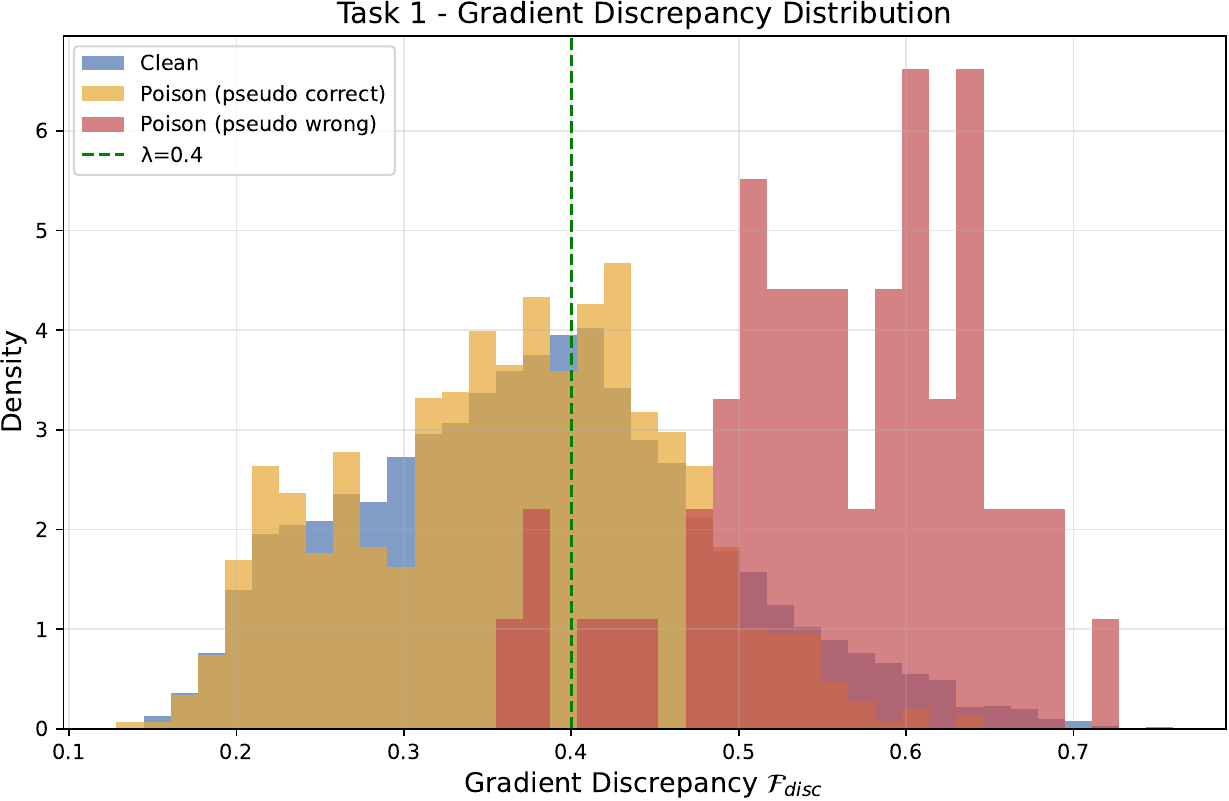}
\end{minipage}}
    \hfill
    \subfloat{%
\begin{minipage}[c]{0.31\textwidth}
\centering
        \includegraphics[width=\textwidth]{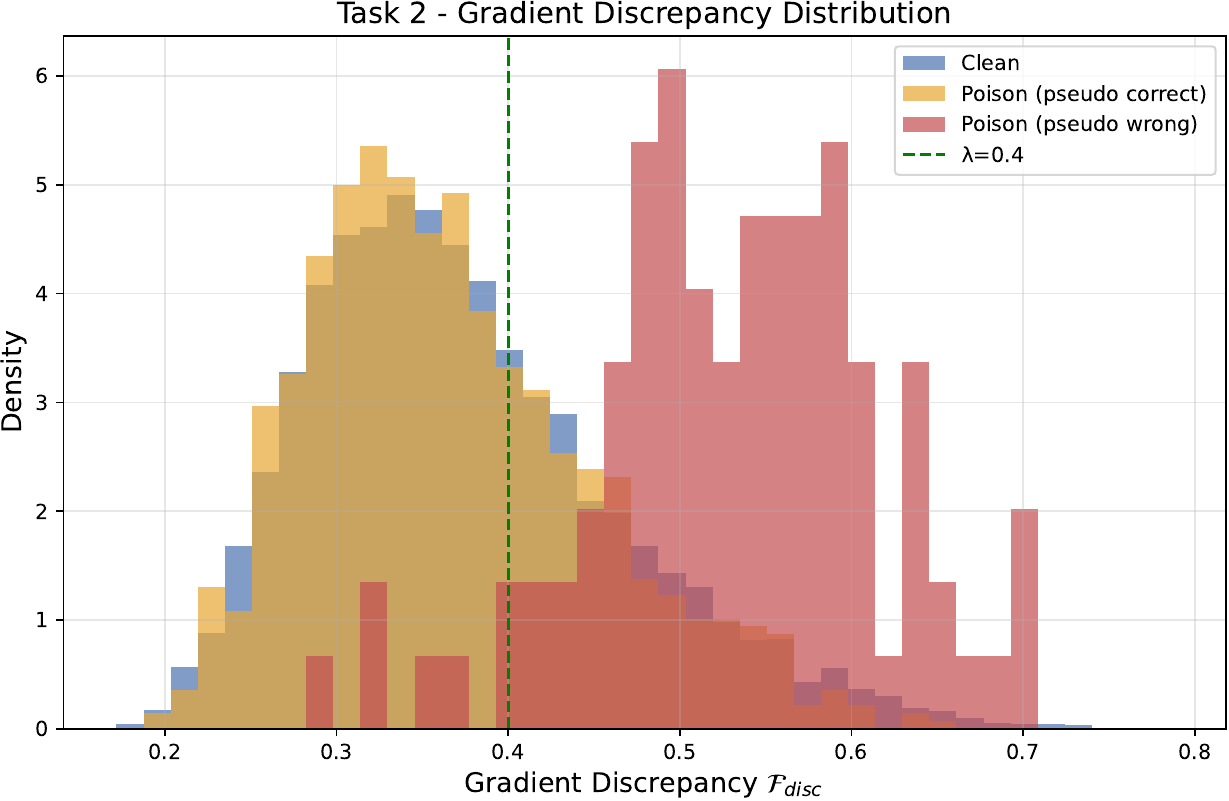}
\end{minipage}}
    
    \par\vspace{6pt}
\begin{center}
    \subfloat{%
\begin{minipage}[c]{0.31\textwidth}
\centering
        \includegraphics[width=\textwidth]{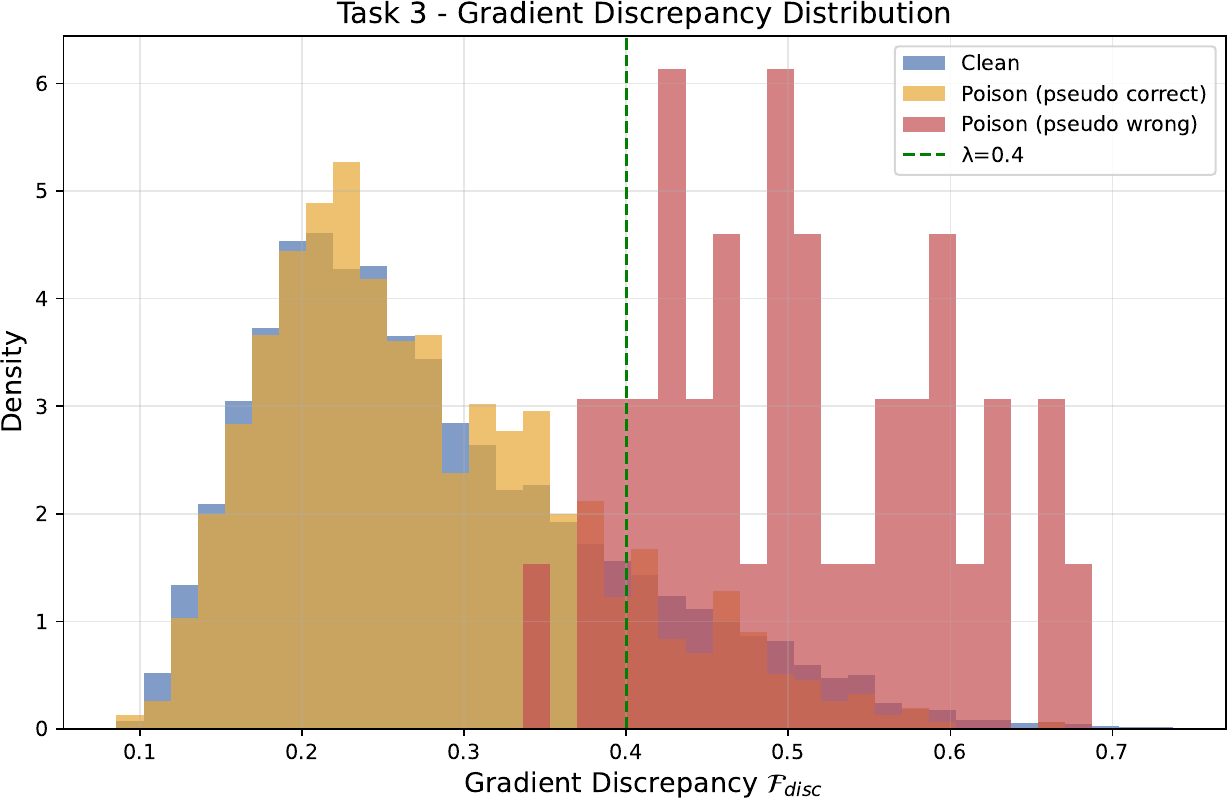}
\end{minipage}}
    \hspace{0.05\textwidth}
    \subfloat{%
\begin{minipage}[c]{0.31\textwidth}
\centering
        \includegraphics[width=\textwidth]{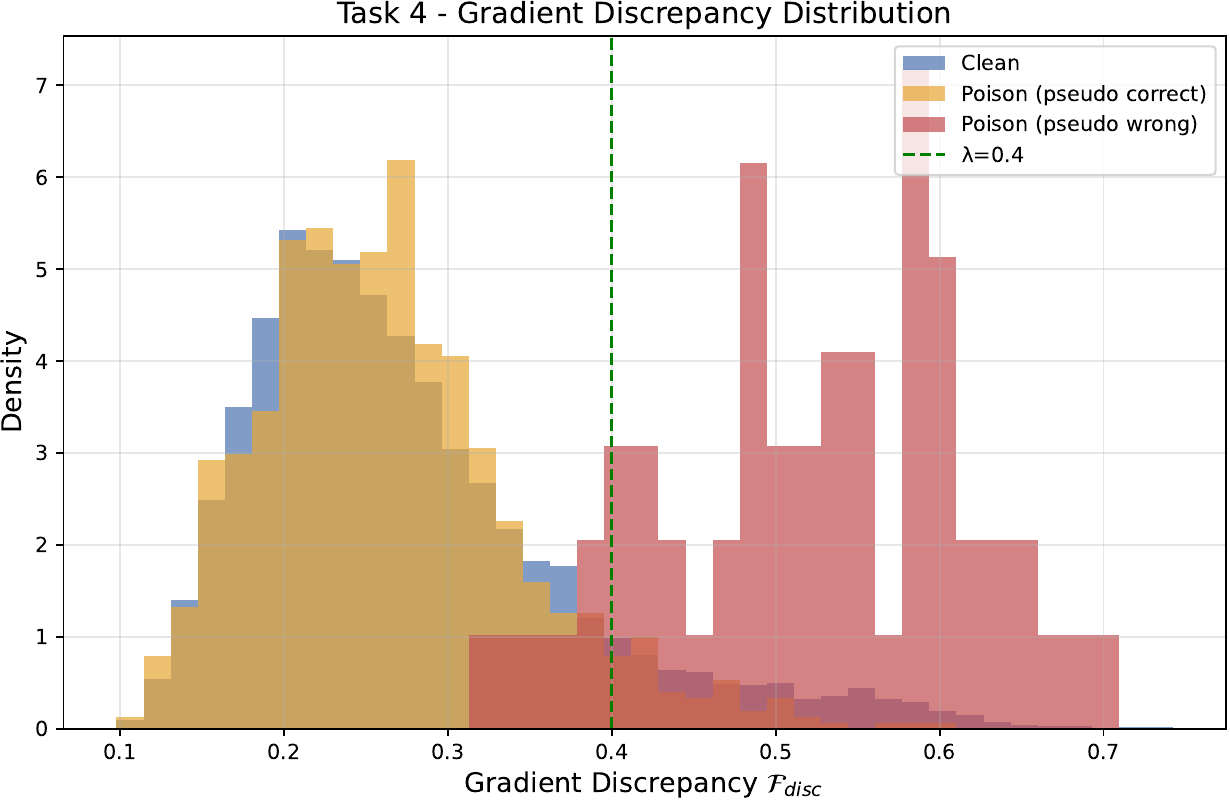}
\end{minipage}}
\end{center}
    \caption{Gradient discrepancy distributions for all five sequential tasks under BadNets attack on CIFAR-10. Clean samples are shown in blue, poisoned samples with correct pseudo-labels in orange, and poisoned samples with wrong pseudo-labels in red. The green dashed line denotes the correction threshold \(\lambda=0.4\).}
    \label{fig:fdisc_hist}
    \vspace{-10pt}
\end{figure*}

\noindent
\textBF{Representative Overhead Profiles.} DER++ uses a fixed architecture, while AIBD provides a comparison with a specialized defense pipeline. Our dynamic architecture adds a lightweight expert for each task while freezing the shared backbone and earlier experts. The reported parameter increment is approximately 0.5M per task, with only the current expert updated. The selected profiles compare this gradual capacity increase with the GPU memory, training time, and inference cost of the two baselines.

\noindent
\textBF{Defense Performance and Computational Cost.} Combined with Table~\ref{tab:exp_combined}, these profiles illustrate the accuracy--ASR trade-off alongside the cost of dynamic expansion. The shared backbone and single selected prediction head keep inference computation close to that of the underlying ViT. Training includes warm-up, correction, and joint optimization; correction evaluates the available experts, and persistent state includes class-reference gradient caches and routing prototypes. These components are included in the scope of resource accounting rather than treating the absence of exemplar replay as zero storage cost.

\subsection{Hyperparameter Sensitivity Analysis}

We evaluate the sensitivity of two key hyperparameters in our proposed defense: the gradient discrepancy threshold $\lambda$ (denoted as \texttt{correction\_threshold}) and the number of warm-up epochs $E_{\text{warm}}$ (denoted as \texttt{warmup\_epochs}). Experiments are conducted on the Seq-CIFAR10 benchmark under BadNets attack with a poisoning rate of 5\%. The clean accuracy (CA, \%) and attack success rate (ASR, \%) are reported in Figure~\ref{fig:hyper_sensitivity}.

\noindent
\textbf{Effect of Gradient Discrepancy Threshold $\lambda$.}
As shown in the left panel of Figure~\ref{fig:hyper_sensitivity}, when $\lambda$ increases from 0.1 to 0.3, the ASR remains near zero (0.02\%--0.00\%) and CA stays above 94\%. This indicates that a small threshold effectively filters out most poisoned samples during the gradient-discrepancy-based selection , because only samples with gradient vectors very close to the clean prototype are accepted. When $\lambda$ exceeds 0.5, the ASR begins to rise (0.31\% at $\lambda=0.5$, 2.70\% at $\lambda=0.7$), while CA declines slightly to 92\%--94\%. This degradation occurs because a larger threshold admits samples with larger gradient discrepancies, some of which are true poisoned samples whose pseudo-labels are incorrect, thereby compromising both robustness and clean accuracy. Notably, at $\lambda=0.6$ and $0.7$, the ASR increases sharply, suggesting that the gradient discrepancy criterion becomes too permissive, admitting samples with weaker class-reference consistency. Within this sweep, the results favor $\lambda \in [0.1, 0.4]$ for the displayed accuracy--ASR trade-off.
Note that in the subsequent warm-up analysis we fix $\lambda = 0.5$; although this value lies slightly outside the strict optimal range, it still yields acceptable ASR ($0.31\%$) and high CA, and importantly it resides at the transition point where the defense transitions from very robust to gradually more permissive. Choosing $\lambda = 0.5$ allows the warm-up study to be conducted in a regime where the effect of $E_{\text{warm}}$ is clearly observable, without being masked by an overly stringent filter that might otherwise limit the dynamic range of the performance metrics.

\noindent
\textbf{Effect of Warm-up Epochs $E_{\text{warm}}$.}
The right panel of Figure~\ref{fig:hyper_sensitivity} depicts the results for $E_{\text{warm}} \in \{1,3,5,7,10,15\}$ with $\lambda=0.5$. The warm-up phase trains the current expert exclusively on the clean set $\mathbf{H}^i$ using $\mathcal{L}_{\text{warm}}$ . A small number of epochs (e.g., 1--3) yields moderate CA ($\sim92\%$) but relatively higher ASR (0.45\% at $E_{\text{warm}}=1$). With 5--10 epochs, CA stabilizes around 93\%--94\% and ASR drops to 0.05\%--0.31\%, indicating that sufficient warm-up enables the expert to learn discriminative class prototypes $\mathbf{S}^{j,i}$  that are robust to residual noise. However, further increasing $E_{\text{warm}}$ to 15 epochs does not improve CA and slightly elevates ASR (0.67\%), likely due to overfitting to the limited clean subset, which reduces the model's ability to generalize to corrected samples during the integration optimization . Thus, $E_{\text{warm}}=5$ or $10$ provides the best balance.
The warm-up results describe the slice at $\lambda=0.5$; interactions with other thresholds are not measured by this one-dimensional sweep.

Overall, the proposed defense is stable within a reasonable range of hyperparameters: CA fluctuates within $\sim$2\% and ASR remains below 1\% for most settings. These results confirm that our BPP and GDBRO mechanisms are not overly sensitive to hyperparameter choices, which is desirable for practical deployment in continual learning under poisoning attacks.

\subsection{Analysis of Gradient Discrepancy Distributions}
\label{sec:gradient_diagnostics}

\textbf{Experimental setup.}
We conduct experiments on the CIFAR-10 dataset under the sequential class-incremental
learning protocol, where the 10 classes are divided into 5 tasks with 2 classes per task.
The backdoor attack adopts the BadNets scheme with a \(3\times 3\) white patch trigger
placed at the bottom-right corner, a target label \(0\), and \textbf{for this diagnostic
experiment we deliberately raise the poisoning rate to \(10\%\)} (instead of the \(5\%\)
used in all other main experiments) to obtain a richer set of poisoned samples for
visualizing the gradient discrepancy distributions.
The defense hyperparameters are set to warm-up epochs \(E_{\text{warm}}=5\) and correction
threshold \(\lambda=0.4\).

\noindent
\textbf{Validation of GDBRO.} The Gradient Discrepancy-Based Robustness Optimization (GDBRO) module plays a central role in our defense framework by identifying and correcting poisoned samples that would otherwise degrade model robustness. As defined in Eq.~\eqref{eq:dis}, the gradient discrepancy \(\mathcal{F}_{\text{disc}}\) measures the cosine distance between the gradient of a candidate sample \(\tilde{\mathbf{x}}\) and the gradient of its pseudo-label class prototype \(\mathbf{S}^{y',s^\star}\) with respect to the classifier parameters. The correction score measures agreement with the class reference. Therefore, a suitable threshold \(\lambda\) can effectively separate clean and correctly corrected samples from mislabeled backdoor ones.

Figure~\ref{fig:fdisc_hist} visualizes the \(\mathcal{F}_{\text{disc}}\) distributions for all five sequential tasks, where samples are grouped into three categories: clean samples, poisoned samples that receive correct pseudo-labels, and poisoned samples that receive wrong pseudo-labels. It is evident that clean samples consistently concentrate at low \(\mathcal{F}_{\text{disc}}\) values, typically below \(0.2\), confirming that their gradient directions are highly aligned with the corresponding class prototypes. Poisoned samples with correct pseudo-labels also exhibit predominantly small discrepancies, only slightly higher than clean ones, because the trigger perturbation is not sufficient to alter the semantic class intended by the attacker. In sharp contrast, poisoned samples whose pseudo-labels are erroneous (e.g., predicting the target class instead of the original ground-truth label) display markedly larger \(\mathcal{F}_{\text{disc}}\) values, forming a distinct distribution shifted to the right. This clear separation demonstrates the effectiveness of GDBRO in distinguishing suspicious backdoor examples from benign ones.

\noindent
\textbf{Justification of \(\lambda=0.4\).} The choice of the hyperparameter \(\lambda\) directly influences the quality of the robustness subset \(\hat{\mathbf{H}}^i\) formed by the threshold rule. A too stringent threshold would exclude many poisoned samples that could otherwise be corrected, while a too liberal threshold risks admitting genuinely corrupted samples. From Figure~\ref{fig:fdisc_hist}, we observe that the majority of clean samples and poison-correct samples lie well below \(0.4\), whereas a substantial portion of poison-wrong samples exceeds this value. Setting \(\lambda=0.4\) thus achieves a favorable trade-off: it retains most clean and correctly pseudo-labeled data while filtering out a large fraction of poisoned samples with highly inconsistent gradient signatures. This is an empirical diagnostic of selective recovery, rather than a guarantee that every accepted pseudo-label is correct.

Notably, Task~0, which contains the target class, exhibits a higher absolute count of poisoned samples, but the discriminative pattern remains consistent with other tasks. The overall similarity of the histograms across tasks further demonstrates the stability of GDBRO under the dynamic expansion regime, making the fixed threshold \(\lambda=0.4\) a robust choice throughout the continual learning process.

\section{Conclusion and Limitation}

In this paper, we investigate a challenging yet practical scenario of continual learning under backdoor attacks (CLUBA), where models must simultaneously preserve previously acquired knowledge, adapt to newly arriving tasks, and resist malicious supervision injected during incremental updates. To address this challenge, we propose a robust dynamic-expansion framework that integrates backdoor purification, selective sample recovery, and robust expert routing into a unified continual learning paradigm. Specifically, the proposed Bi-Prototype Purification (BPP) mechanism identifies suspicious samples by exploiting semantic inconsistencies in feature space, while Gradient Discrepancy-based Robustness Optimization (GDBRO) further evaluates and selectively recovers informative samples through gradient-level consistency. Moreover, Robust Feature Consistency-based Expert Selection (RFCBES) improves inference-time robustness by constructing reliable class prototypes for expert routing.

Extensive experiments on multiple continual backdoor benchmarks demonstrate that our framework achieves a favorable balance among stability, plasticity, and security. Unlike conventional continual learning approaches that mainly focus on preventing catastrophic forgetting, our method explicitly considers the risk of absorbing corrupted supervision during sequential learning. The results validate that combining task-local purification with selective recovery provides an effective solution for maintaining model adaptability while reducing vulnerability to persistent backdoor attacks.

Despite these promising results, several limitations remain. First, although we evaluate our framework against representative continual backdoor attacks, the current study does not exhaustively cover all possible attack strategies, especially adaptive attackers that may intentionally manipulate feature distributions to evade purification or gradient-based detection. A broader evaluation involving more diverse and adaptive attack scenarios is an important direction for future research. Second, the effectiveness of BPP relies on the assumption that poisoned samples exhibit sufficient semantic deviation from clean samples in the learned representation space. When attackers design stealthier triggers that preserve strong semantic alignment with clean samples, purification performance may degrade. Developing attack-agnostic purification strategies with weaker distribution assumptions remains an open problem. Third, the current GDBRO module adopts a fixed discrepancy threshold for sample recovery, which may require additional adaptation when task distributions or attack intensities vary significantly. Future work could explore adaptive threshold estimation or meta-learning-based calibration strategies to improve scalability in more dynamic environments.

Finally, our experiments mainly focus on supervised image classification benchmarks with controlled task sequences. Extending the proposed framework to large-scale real-world continual learning systems, including multimodal models, streaming applications, and open-world scenarios with unknown task boundaries, represents another promising research direction. We believe that developing secure and adaptive continual learners capable of continuously acquiring knowledge while resisting evolving threats will remain an important challenge for future machine learning systems.

\bibliographystyle{IEEEtran}
\bibliography{main}

\vspace{-40pt}
\begin{IEEEbiography}
[{\includegraphics[width=1in,height=1.25in,clip,keepaspectratio]{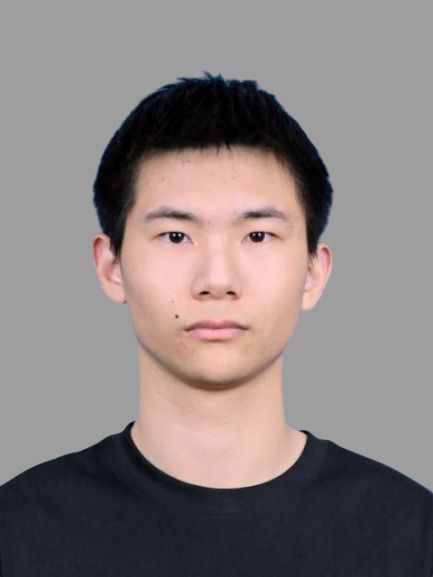}}]{Keyu Lin} is currently pursuing the master degree at University of Electronic Science and Technology of China (UESTC). He received the bachelor degree from Southwest Jiaotong University, China, in 2025. His research interests include continual learning and deep learning.
\end{IEEEbiography}
\vspace{-40pt}
\begin{IEEEbiography}
[{\includegraphics[width=1in,height=1.25in,clip,keepaspectratio]{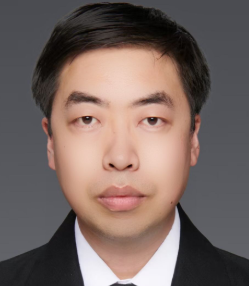}}]{Fei Ye} is currently a full professor at University of Electronic Science and Technology of China (UESTC). He got the PhD degree from University of York. He received the bachelor degree from Chengdu University of Technology, China, in 2014 and the master degree in computer science and technology from Southwest Jiaotong University, China, in 2018. His research topics includes deep generative image models, lifelong learning and mixture models.
\end{IEEEbiography}
\vspace{-30pt}
\begin{IEEEbiography}
[{\includegraphics[width=1in,height=1.25in,clip,keepaspectratio]{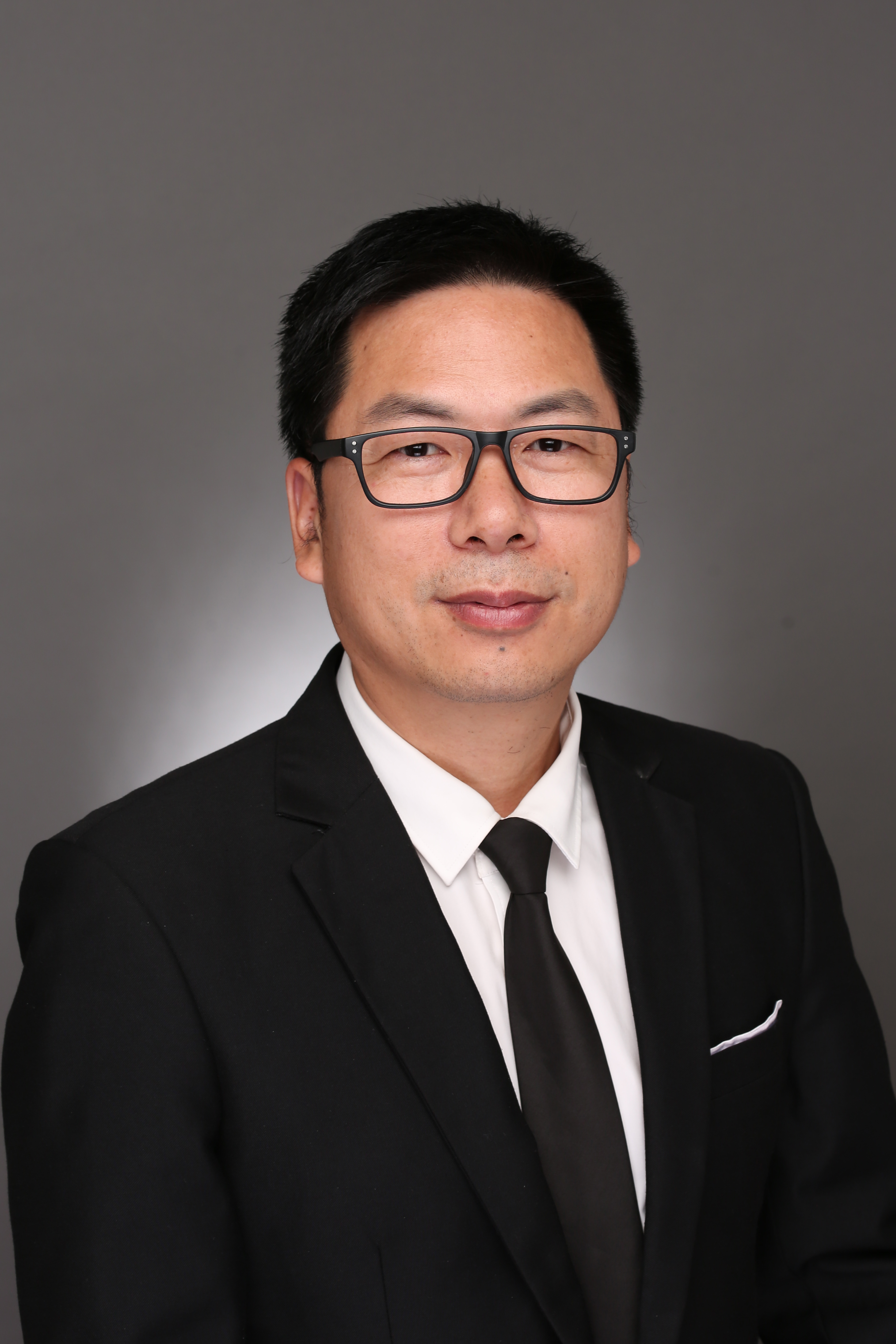}}]{Qihe Liu} received the B.S. degree in mathematics from Sichuan Normal University, Chengdu, China, in 1995, the M.S. degree in mathematics from Sichuan University, Chengdu, China, in 1998, and the Ph.D. degree in computer application technology from the University of Electronic Science and Technology of China (UESTC), Chengdu, China, in 2005. Since 1998, he has been with the School of Computer Science and Engineering, UESTC, where he is currently an Associate Professor. 
\end{IEEEbiography}
\vspace{-40pt}
\begin{IEEEbiography}
[{\includegraphics[width=1in,height=1.25in,clip,keepaspectratio]{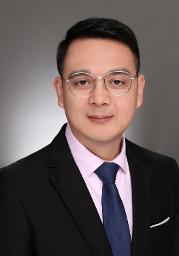}}]{Shijie Zhou} (Member, IEEE) received the Ph.D. degree in computer application technology from the University of Electronic Science and Technology of China (UESTC), Chengdu, China, in June 2004, where he has since been with the School of Computer Science and Engineering. He was a Visiting Scholar at the University of Hong Kong from September 2006 to April 2007 and at Purdue University from January 2009 to January 2010. 
\end{IEEEbiography}
\vspace{-460pt}
\begin{IEEEbiography}
[{\includegraphics[width=1in,height=1.25in,clip,keepaspectratio]{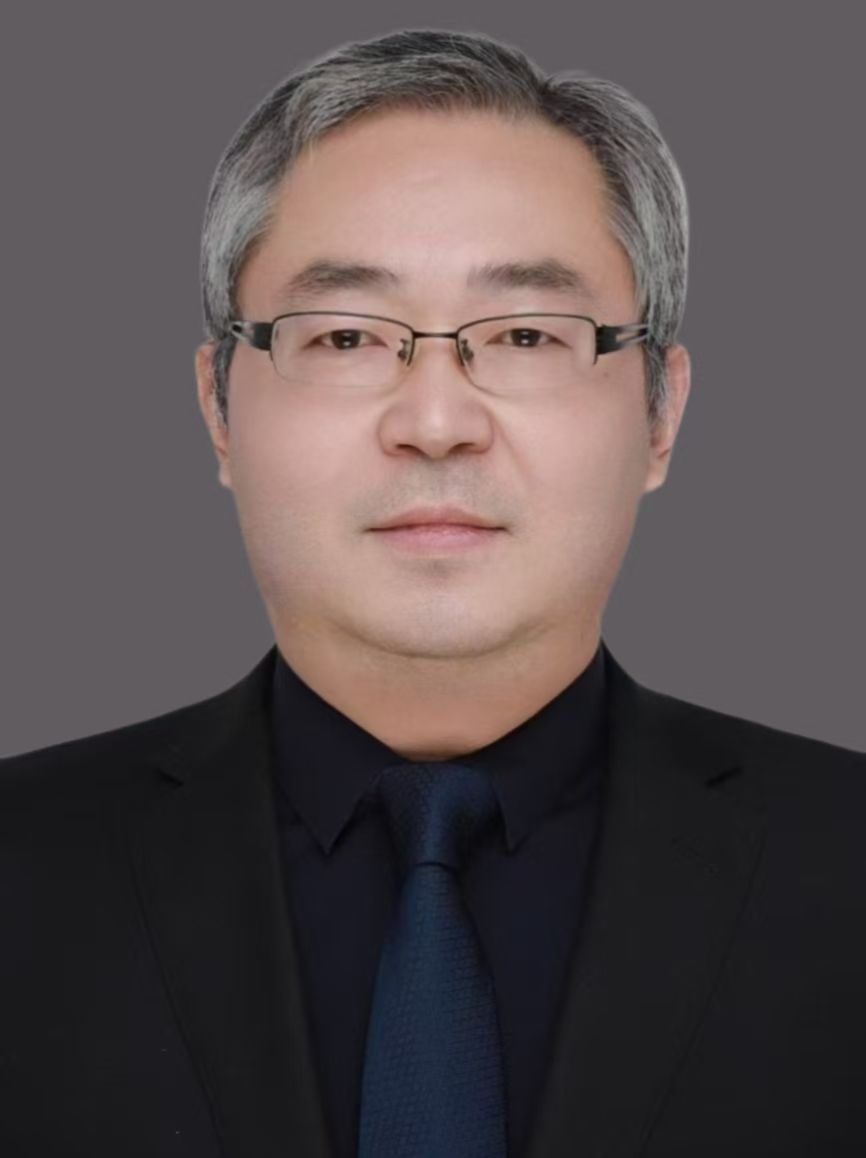}}]{Jiguo Yu} (Fellow, IEEE) received the B.S. and M.S. degrees from Qufu Normal University, Qufu, China, and the Ph.D. degree from Shandong University, Jinan, China. He is currently a Professor with the School of Information and Software Engineering, University of Electronic Science and Technology of China (UESTC), Chengdu, China. He was previously a Professor and Dean with the School of Computer Science and Technology, Qilu University of Technology, Jinan, China, and a Professor with Qufu Normal University. 
\end{IEEEbiography}

\end{document}